\pdfoutput=1
\def\modePreprint{}

\ifx\modeSubmission\undefined
  \ifx\modePreprint\undefined
    \ifx\modeFinal\undefined
      \ifx\modeWebsite\undefined
        \ifx\modeQuals\undefined
          \def\modeDev{}
        \fi
      \fi
    \fi
  \fi
\fi

\newcommand{\ifModeWebsite}[1]{\ifdefined\modeWebsite#1\fi}
\newcommand{\ifModeWebsiteOff}[1]{\ifx\modeWebsite\undefined#1\fi}
\newcommand{\ifModeQuals}[1]{\ifdefined\modeQuals#1\fi}
\newcommand{\ifModeQualsOff}[1]{\ifx\modeQuals\undefined#1\fi}

\ifdefined\modeSubmission
  \documentclass[sigconf,anonymous,review]{acmart}
  \newcommand{\hidenotes}{}
\fi
\ifdefined\modePreprint

  \documentclass[sigconf]{acmart}
  \newcommand{\hidenotes}{}
\fi
\ifdefined\modeFinal
  \documentclass[sigconf]{acmart}
  \newcommand{\hidenotes}{}
\fi
\ifdefined\modeQuals
  \documentclass[acmsmall,nonacm]{acmart}
  \newcommand{\hidenotes}{}
\fi
\ifdefined\modeWebsite
  \documentclass{article}

  \newcommand{\hidenotes}{}

  \usepackage{latexml}
  \usepackage{graphicx}

  \usepackage[T1]{fontenc}

  \newcommand{\citep}[1]{\cite{#1}}

  \newcommand{\affiliation}[1]{#1}
  \newcommand{\institution}[1]{#1}
  
  \newcommand{\city}[1]{#1}
  \newcommand{\state}[1]{#1}
  \newcommand{\country}[1]{#1}
  \newcommand{\email}[1]{#1}

  \newcommand\Description[2][]{}

  \usepackage{comment}
  \includecomment{anonsuppress}

  \newcommand\acksname{Acknowledgments}
  \specialcomment{acks}{%
    \begingroup
    \section*{\acksname}
    \phantomsection\addcontentsline{toc}{section}{\acksname}
  }{%
    \endgroup
  }
\fi
\ifdefined\modeDev

  \documentclass[sigconf]{acmart}
\fi

\AtBeginDocument{%
  \providecommand\BibTeX{{%
    \normalfont B\kern-0.5em{\scshape i\kern-0.25em b}\kern-0.8em\TeX}}}

\ifModeWebsiteOff{

\copyrightyear{2022}
\acmYear{2022}
\setcopyright{rightsretained}
\acmConference[GECCO '22]{Genetic and Evolutionary Computation Conference}{July 9--13, 2022}{Boston, MA, USA}
\acmBooktitle{Genetic and Evolutionary Computation Conference (GECCO '22), July 9--13, 2022, Boston, MA, USA}
\acmDOI{10.1145/3512290.3528705}
\acmISBN{978-1-4503-9237-2/22/07}

}

\usepackage{amsmath,amsfonts,bm}

\def\eqref#1{equation~\ref{#1}}

\def\1{\bm{1}}

\def\vmu{{\bm{\mu}}}
\def\vtheta{{\bm{\theta}}}

\def\vc{{\bm{c}}}

\def\vm{{\bm{m}}}

\def\vp{{\bm{p}}}

\def\vx{{\bm{x}}}

\def\mI{{\bm{I}}}

\def\mSigma{{\bm{\Sigma}}}

\DeclareMathAlphabet{\mathsfit}{\encodingdefault}{\sfdefault}{m}{sl}
\SetMathAlphabet{\mathsfit}{bold}{\encodingdefault}{\sfdefault}{bx}{n}

\def\gN{{\mathcal{N}}}

\newcommand{\E}{\mathbb{E}}

\newcommand{\R}{\mathbb{R}}

\DeclareMathOperator*{\argmin}{arg\,min}

\newcommand \cmaes {\mbox{CMA}-\mbox{ES}}
\newcommand \cmame {\mbox{CMA}-\mbox{ME}}
\newcommand \cmamega {\mbox{CMA}-\mbox{MEGA}}
\newcommand \cmamegaes {\mbox{CMA}-\mbox{MEGA} (ES)}
\newcommand \cmamegatdes {\mbox{CMA}-\mbox{MEGA} (TD3, ES)}
\newcommand \mapelites {\mbox{MAP}-\mbox{Elites}}
\newcommand \mees {\mbox{ME}-\mbox{ES}}
\newcommand \nses {\mbox{NS}-\mbox{ES}}
\newcommand \nsres {\mbox{NSR}-\mbox{ES}}
\newcommand \nsraes {\mbox{NSRA}-\mbox{ES}}
\newcommand \openaies {\mbox{OpenAI}-\mbox{ES}}
\newcommand \ogmapelites {\mbox{OG}-\mbox{MAP}-\mbox{Elites}}

\newcommand \pgame {\mbox{PGA}-\mbox{MAP}-\mbox{Elites}}

\newcommand \qdrl {\mbox{QD}-\mbox{RL}}

\usepackage{xcolor}

\newcommand{\xxnote}[3]{}
\ifx\hidenotes\undefined
  \renewcommand{\xxnote}[3]{\color{#2}{(#1: #3)}}
\fi

\newcommand{\hlc}[2][yellow]{{\sethlcolor{#1} \hl{#2}}}

\definecolor{lightgrey}{rgb}{0.9, 0.9, 0.9}

\newcommand{\eref}[1]{Eq. \ref{#1}}
\newcommand{\sref}[1]{Sec. \ref{#1}}
\newcommand{\apref}[1]{Appendix \ref{#1}}
\newcommand{\fref}[1]{Fig. \ref{#1}}
\newcommand{\tref}[1]{Table \ref{#1}}
\newcommand{\aref}[1]{Algorithm \ref{#1}}

\def\vnabla{{\bm{\nabla}}}
\def\vphi{{\bm{\phi}}}
\def\vepsilon{{\bm{\epsilon}}}

\usepackage{booktabs}
\usepackage{multirow}
\usepackage{array}
\newcolumntype{L}[1]
  {>{\raggedright\let\newline\\\arraybackslash\hspace{0pt}}m{#1}}
\newcolumntype{C}[1]
  {>{\centering\let\newline\\\arraybackslash\hspace{0pt}}m{#1}}
\newcolumntype{R}[1]
  {>{\raggedleft\let\newline\\\arraybackslash\hspace{0pt}}m{#1}}

\usepackage{hyperref}
\usepackage{soul} %
\usepackage{caption}
\usepackage{float}
\usepackage[ruled,commentsnumbered,linesnumbered]{algorithm2e} %
\usepackage{lscape}
\DontPrintSemicolon

\begin{document}

\title{Approximating Gradients for Differentiable Quality Diversity in Reinforcement Learning}

\ifdefined\modeWebsite

\author{\href{https://btjanaka.net}{Bryon Tjanaka}\\
        University of Southern California\\
        \href{mailto:tjanaka@usc.edu}{\lxAddClass{ltx2_author_email}tjanaka@usc.edu}}
\author{\href{https://scholar.google.com/citations?user=RqSvzikAAAAJ}{Matthew C. Fontaine}\\
        University of Southern California\\
        \href{mailto:mfontain@usc.edu}{\lxAddClass{ltx2_author_email}mfontain@usc.edu}}
\author{\href{http://julian.togelius.com}{Julian Togelius}\\
        New York University\\
        \href{mailto:julian@togelius.com}{\lxAddClass{ltx2_author_email}julian@togelius.com}}
\author{\href{https://stefanosnikolaidis.net}{Stefanos Nikolaidis}\\
        University of Southern California\\
        \href{mailto:nikolaid@usc.edu}{\lxAddClass{ltx2_author_email}nikolaid@usc.edu}}

\else

  \ifdefined\modeQuals

\author{Bryon Tjanaka}
\affiliation{%
  \institution{University of Southern California}
  \city{Los Angeles}
 \state{California}
  \country{USA}}
\email{tjanaka@usc.edu}

  \else

\author{Bryon Tjanaka}
\affiliation{%
  \institution{University of Southern California}
  \city{Los Angeles}
 \state{California}
  \country{USA}}
\email{tjanaka@usc.edu}

\author{Matthew C. Fontaine}
\affiliation{%
  \institution{University of Southern California}
  \city{Los Angeles}
 \state{California}
  \country{USA}}
\email{mfontain@usc.edu}

\author{Julian Togelius}
\affiliation{%
  \institution{New York University}
  \city{Brooklyn}
 \state{New York}
  \country{USA}}
\email{julian@togelius.com}

\author{Stefanos Nikolaidis}
\affiliation{%
  \institution{University of Southern California}
  \city{Los Angeles}
 \state{California}
  \country{USA}}
\email{nikolaid@usc.edu}

  \fi

\fi

\begin{abstract}

  Consider the problem of training robustly capable agents. One approach is to generate a diverse collection of agent polices. Training can then be viewed as a quality diversity (QD) optimization problem, where we search for a collection of performant policies that are diverse with respect to quantified behavior. Recent work shows that differentiable quality diversity (DQD) algorithms greatly accelerate QD optimization when exact gradients are available. However, agent policies typically assume that the environment is not differentiable. To apply DQD algorithms to training agent policies, we must approximate gradients for performance and behavior. We propose two variants of the current state-of-the-art DQD algorithm that compute gradients via approximation methods common in reinforcement learning (RL). We evaluate our approach on four simulated locomotion tasks. One variant achieves results comparable to the current state-of-the-art in combining QD and RL, while the other performs comparably in two locomotion tasks. These results provide insight into the limitations of current DQD algorithms in domains where gradients must be approximated. Source code is available at \url{https://github.com/icaros-usc/dqd-rl}

\end{abstract}

\ifx\modeWebsite\undefined
\fi

\ifx\modeWebsite\undefined
\fi

\maketitle

\section{Introduction}

\begin{figure}[t]
\ifModeWebsite{\lxAddClass{mount_diagram}}
\begin{center}
  \ifdefined\modeJournal
    \includegraphics[width=0.8\linewidth]{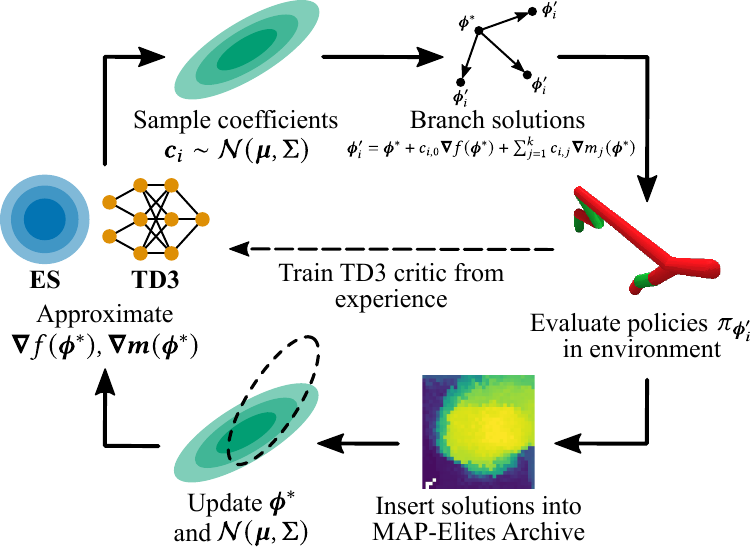}
  \else
    \ifdefined\modeIclr
      \includegraphics[width=0.5\linewidth]{imgs/diagram/diagram.pdf}
    \else
      \includegraphics[width=0.95\linewidth]{imgs/diagram/diagram.pdf}
    \fi
  \fi
\end{center}
\caption{We develop two RL variants of the \cmamega{} algorithm. Similar to \cmamega{}, the variants sample gradient coefficients $\vc$ and branch around a solution point $\vphi^*$. We evaluate each branched solution $\vphi'_i$ as part of a policy $\pi_{\vphi'_i}$ and insert $\vphi'_i$ into the archive.
We then update $\vphi^*$ and $\mathcal{N}(\vmu,\mSigma)$ to maximize archive improvement. Our RL variants differ from \cmamega{} by approximating gradients with ES and TD3, since exact gradients are unavailable in RL settings.}
\label{fig:diagram}
\Description[Fully described in the text]{}
\end{figure}

\ifModeQuals{\begin{figure}[t]
\ifModeWebsite{\lxAddClass{mount_frames}}
\begin{center}
  \ifdefined\modeIclr
    \includegraphics[width=0.5\linewidth]{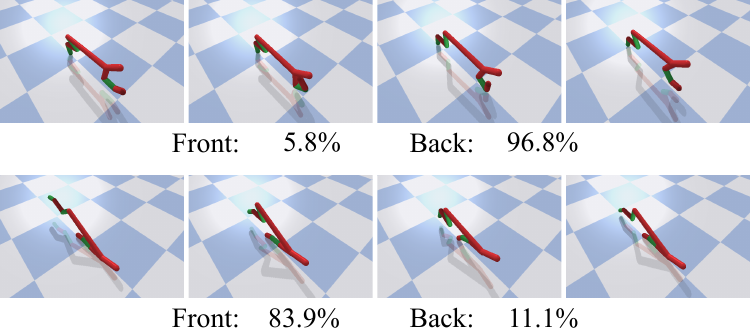}
  \else
    \includegraphics[width=\linewidth]{imgs/frames/frames.pdf}
  \fi
\end{center}
\ifdefined\modeWebsite
  \caption{A half-cheetah agent executing two walking policies. On the left, the agent walks on its back foot while tapping the ground with its front foot. On the right, the agent walks on its front foot while jerking its back foot. Values on each video show the percentage of time each foot contacts the ground (each foot is measured individually, so values do not sum to 100\%). With these policies, the agent could continue walking even if one foot is damaged.}
\else
  \caption{A half-cheetah agent executing two walking policies. In the top row, the agent walks on its back foot while tapping the ground with its front foot. In the bottom row, the agent walks on its front foot while jerking its back foot. Values below each row show the percentage of time each foot contacts the ground (each foot is measured individually, so values do not sum to 100\%). With these policies, the agent could continue walking even if one foot is damaged.}
\fi
\label{fig:frames}
\Description[Fully described in the text]{}
\end{figure}
}

We focus on the problem of extending differentiable quality diversity (DQD) to reinforcement learning (RL) domains. We propose to approximate gradients for the objective and measure functions, resulting in two variants of the DQD algorithm \cmamega{}~\citep{fontaine2021dqd}.

Consider a half-cheetah agent (\fref{fig:frames}) trained for locomotion, where the agent must continue walking forward even when one foot is damaged. If we frame this challenge as an RL problem, two approaches to design a robustly capable agent would be to (1) design a reward function and (2) apply domain randomization \citep{tobin2017domain,peng2018domain}. However, prior work \citep{rlblogpost,faultyrewards} suggests that designing such a reward function is difficult, while domain randomization may require manually selecting hundreds of environment parameters \citep{peng2018domain,openai2019rubiks}.

As an alternative approach, consider that we have intuition on what behaviors would be useful for adapting to damage. For instance, we can \textit{measure} how often each foot is used during training, and we can pre-train a collection of policies that are diverse in how the agent uses its feet. When one of the agent's feet is damaged during deployment, the agent can adapt to the damage by selecting a policy that did not move the damaged foot during training \citep{cully2015, colas2020scaling}.

\ifModeWebsite{\begin{figure}[t]
\ifModeWebsite{\lxAddClass{mount_frames}}
\begin{center}
  \ifdefined\modeIclr
    \includegraphics[width=0.5\linewidth]{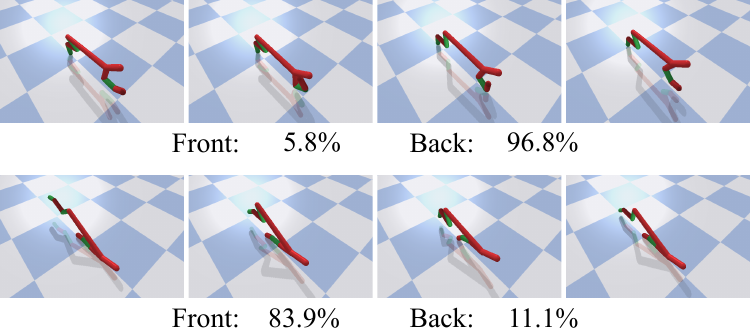}
  \else
    \includegraphics[width=\linewidth]{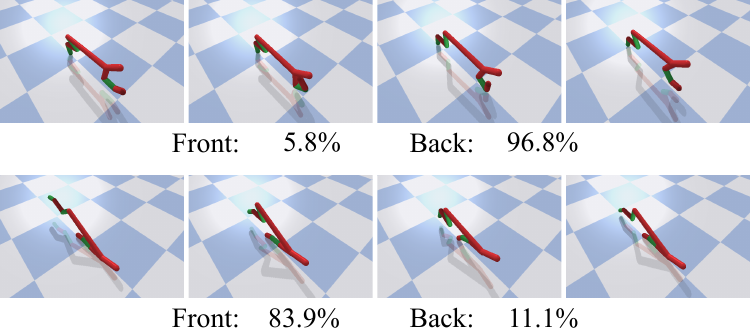}
  \fi
\end{center}
\ifdefined\modeWebsite
  \caption{A half-cheetah agent executing two walking policies. On the left, the agent walks on its back foot while tapping the ground with its front foot. On the right, the agent walks on its front foot while jerking its back foot. Values on each video show the percentage of time each foot contacts the ground (each foot is measured individually, so values do not sum to 100\%). With these policies, the agent could continue walking even if one foot is damaged.}
\else
  \caption{A half-cheetah agent executing two walking policies. In the top row, the agent walks on its back foot while tapping the ground with its front foot. In the bottom row, the agent walks on its front foot while jerking its back foot. Values below each row show the percentage of time each foot contacts the ground (each foot is measured individually, so values do not sum to 100\%). With these policies, the agent could continue walking even if one foot is damaged.}
\fi
\label{fig:frames}
\Description[Fully described in the text]{}
\end{figure}
}

Pre-training such a collection of policies may be viewed as a quality diversity (QD) optimization problem \citep{pugh2016qd,cully2015,mouret2015illuminating, colas2020scaling}. Formally, QD assumes an objective function $f$ and one or more measure functions $\vm$. The goal of QD is to find solutions satisfying all output combinations of $\vm$, i.e. moving different combinations of feet, while maximizing each solution's $f$, i.e. walking forward quickly. Most QD algorithms treat $f$ and $\vm$ as black boxes, but recent work \citep{fontaine2021dqd} proposes differentiable quality diversity (DQD), which assumes $f$ and $\vm$ are differentiable functions with exact gradient information. 
 QD algorithms have been applied to procedural content generation \citep{gravina2019procedural},  robotics \citep{cully2015, mouret2015illuminating}, aerodynamic shape design \citep{gaier2018dataefficient}, and scenario generation in human-robot interaction \cite{fontaine2021quality,fontaine2021importance}.  

The recently proposed DQD algorithm \cmamega{} \citep{fontaine2021dqd} outperforms QD algorithms by orders of magnitude when exact gradients are available, such as when searching the latent space of a generative model. However, RL problems like the half-cheetah lack these gradients because the environment is typically non-differentiable, thus limiting the applicability of DQD. To address this limitation, we draw inspiration from how evolution strategies (ES) \citep{akimoto2010,wierstra2014nes,salimans2017evolution,mania2018ars} and deep RL actor-critic methods \citep{schulman2015trpo,schulman2017ppo,lillicrap2016ddpg,fujimoto2018td3} optimize a reward objective by approximating gradients for gradient descent. \textit{Our key insight is to approximate objective and measure gradients for DQD algorithms by adapting ES and actor-critic methods.}

Our work makes three contributions. \textbf{(1)} We formalize the problem of quality diversity for reinforcement learning (\qdrl{}) and reduce it to an instance of DQD. \textbf{(2)} We develop two \qdrl{} variants of the DQD algorithm \cmamega{}, where each algorithm approximates objective and measure gradients with a different combination of ES and actor-critic methods. \textbf{(3)} We benchmark our variants on four PyBullet locomotion tasks from QDGym \citep{benelot2018, qdgym}. One variant performs comparably (in terms of QD score; \sref{sec:metrics}) to the state-of-the-art \pgame{} \citep{nilsson2021pga} in two tasks. The other variant achieves comparable QD score with \pgame{} in all tasks\footnote{We note that the performance of the CMA-MEGA is worse than \pgame{} in two of the tasks, albeit within variance. We consider it likely that additional runs would result in \pgame{} performing significantly better in these tasks. We leave further evaluation for future work.} but is less efficient than \pgame{} in two tasks.  
\begin{anonsuppress}
\end{anonsuppress}

\ifModeQualsOff{\ifModeWebsiteOff{\begin{figure}[t]
\ifModeWebsite{\lxAddClass{mount_frames}}
\begin{center}
  \ifdefined\modeIclr
    \includegraphics[width=0.5\linewidth]{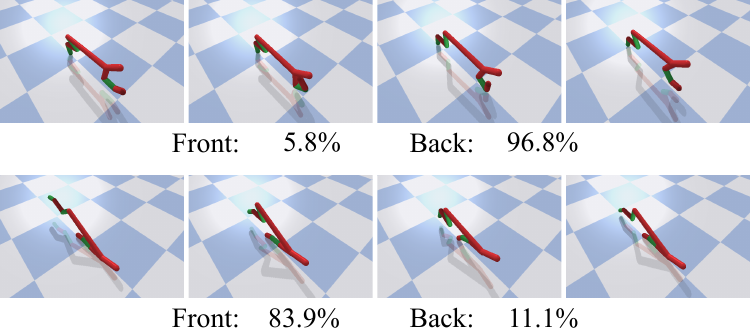}
  \else
    \includegraphics[width=\linewidth]{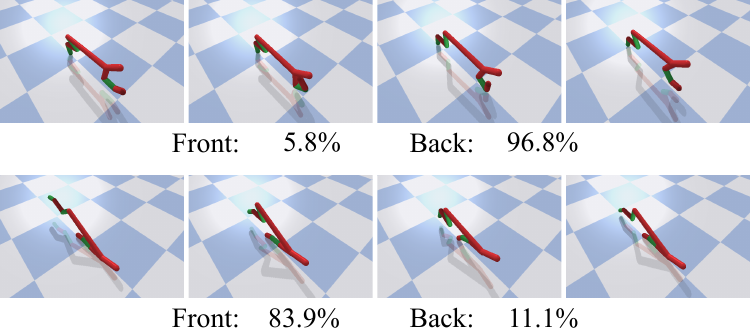}
  \fi
\end{center}
\ifdefined\modeWebsite
  \caption{A half-cheetah agent executing two walking policies. On the left, the agent walks on its back foot while tapping the ground with its front foot. On the right, the agent walks on its front foot while jerking its back foot. Values on each video show the percentage of time each foot contacts the ground (each foot is measured individually, so values do not sum to 100\%). With these policies, the agent could continue walking even if one foot is damaged.}
\else
  \caption{A half-cheetah agent executing two walking policies. In the top row, the agent walks on its back foot while tapping the ground with its front foot. In the bottom row, the agent walks on its front foot while jerking its back foot. Values below each row show the percentage of time each foot contacts the ground (each foot is measured individually, so values do not sum to 100\%). With these policies, the agent could continue walking even if one foot is damaged.}
\fi
\label{fig:frames}
\Description[Fully described in the text]{}
\end{figure}
}}

These results contrast with prior work \citep{fontaine2021dqd} where \cmamega{} vastly outperforms a DQD algorithm inspired by \pgame{} on benchmark functions where gradient information is available. Overall, we shed light on the limitations of \cmamega{} in QD domains where the main challenge comes from optimizing the objective rather than from exploring measure space. At the same time, since we decouple gradient estimates from QD optimization, our work opens a path for future research that would benefit from independent improvements to either DQD or RL.

\section{Problem Statement} \label{sec:problem}

\subsection{Quality Diversity (QD)}

We adopt the definition of QD from prior work \citep{fontaine2021dqd}. For a solution vector $\vphi \in \R^n$, QD considers an objective function $f(\vphi)$ and $k$ measures\footnote{Prior work refers to measure function outputs as ``behavior characteristics,'' ``behavior descriptors,'' or ``feature descriptors.''} $m_i(\vphi) \in \R$ (for $i\in 1..k$) or, as a joint measure, $\vm(\vphi) \in \R^k$. These measures form a $k$-dimensional measure space $\mathcal{X}$. For every $\vx \in \mathcal{X}$, the QD objective is to find solution $\vphi$ such that $\vm(\vphi) = \vx$ and $f(\vphi)$ is maximized. Since $\mathcal{X}$ is continuous, it would require infinite memory to solve the QD problem, so algorithms in the \mapelites{} family \citep{mouret2015illuminating, cully2015} discretize $\mathcal{X}$ by forming a tesselation $\mathcal{Y}$ consisting of $M$ cells. Thus, we relax the QD problem to one of searching for an \textit{archive} $\mathcal{A}$ consisting of $M$ \textit{elites} $\vphi_i$, one for each cell in $\mathcal{Y}$. Then, the QD objective is to maximize the performance $f(\vphi_i)$ of all elites:
\begin{align}
  \max_{\vphi_{1..M}} \sum_{i=1}^M f(\vphi_i) \label{eq:qdobjective}
\end{align}

\subsubsection{Differentiable Quality Diversity (DQD)}

In DQD, we assume $f$ and $\vm$ are first-order differentiable. We denote the objective gradient as $\vnabla f(\vphi)$, or abbreviated as $\vnabla f$, and the measure gradients as $\vnabla \vm(\vphi)$ or $\vnabla {\vm}$.

\subsection{Quality Diversity for Reinforcement Learning (\qdrl{})} \label{sec:qdrl}

We define \qdrl{} as an instance of the QD problem in which each solution $\vphi$ parameterizes an RL policy $\pi_\vphi$. Then, the objective $f(\vphi)$ is the \textit{expected discounted return} of $\pi_\vphi$, and the measures $\vm(\vphi)$ are functions of $\pi_\vphi$. Formally, drawing on the Markov Decision Process (MDP) formulation \citep{Sutton2018}, we represent \qdrl{} as a tuple $(\mathcal{S}, \mathcal{U}, p, r, \gamma, \vm)$. On discrete timesteps $t$ in an episode of interaction, an agent observes state $s \in \mathcal{S}$ and takes action $a \in \mathcal{U}$ according to a policy $\pi_\vphi(a | s)$. The agent then receives scalar reward $r(s, a)$ and observes next state $s' \in \mathcal{S}$ according to $s' \sim p(\cdot | s, a)$. Each episode thus has a trajectory $\xi = \{s_0, a_0, s_1, a_1,.., s_T\}$, where $T$ is the number of timesteps in the episode, and the probability that policy $\pi_\vphi$ takes trajectory $\xi$ is
\ifdefined\modeWebsite
$p_\vphi(\xi) = p(s_0) \prod_{t=0}^{T-1} \pi_\vphi(a_t|s_t)$ $p(s_{t+1} | s_t,a_t)$.
\else
$p_\vphi(\xi) = p(s_0) \prod_{t=0}^{T-1} \pi_\vphi(a_t|s_t) p(s_{t+1} | s_t,a_t)$.
\fi
Now, we define the \textit{expected discounted return} of policy $\pi_\vphi$ as
\begin{align}
    f(\vphi)=\E_{\xi \sim p_\vphi} \left[ \sum_{t=0}^T \gamma^t r(s_t, a_t) \right]
    \label{eq:return}
\end{align}
where the discount factor $\gamma \in (0,1)$ trades off between short- and long-term rewards. Finally, we quantify the behavior of policy $\pi_\vphi$ via a $k$-dimensional measure function $\vm(\vphi)$.

\subsubsection{\qdrl{} as an instance of DQD}

We reduce \qdrl{} to a DQD problem. Since the exact gradients $\vnabla f$ and $\vnabla \vm$ usually do not exist in \qdrl{}, we must instead approximate them.

\section{Background}

\subsection{Single-Objective Reinforcement Learning}

We review algorithms which train a policy to maximize a single objective, i.e. $f(\vphi)$ in \eref{eq:return}, with the goal of applying these algorithms' gradient approximations to DQD in \sref{sec:cmamegaapprox}.

\subsubsection{Evolution strategies (ES)}

ES \citep{beyer2002} is a class of evolutionary algorithms which optimizes the objective by sampling a population of solutions and moving the population towards areas of higher performance. Natural Evolution Strategies (NES) \citep{wierstra2014nes,wierstra2008} is a type of ES which updates the sampling distribution of solutions by taking steps on distribution parameters in the direction of the natural gradient \citep{amari1998}. For example, with a Gaussian sampling distribution, each iteration of an NES would compute natural gradients to update the mean $\vmu$ and covariance $\mSigma$.

We consider an NES-inspired algorithm \citep{salimans2017evolution} which has demonstrated success in RL domains. This algorithm, which we refer to as \openaies{}, samples $\lambda_{es}$ solutions from an isotropic Gaussian but only computes a gradient step for the mean $\vphi$. Each solution sampled by \openaies{} is represented as $\vphi + \sigma \vepsilon_i$, where $\sigma$ is the fixed standard deviation of the Gaussian and $\vepsilon_i \sim \gN(\mathbf{0}, \mI)$. Once these solutions are evaluated, \openaies{} estimates the gradient as
\begin{align}
  \vnabla f(\vphi) \approx \frac{1}{\lambda_{es}\sigma} \sum_{i=1}^{\lambda_{es}} f(\vphi + \sigma \vepsilon_i) \vepsilon_i \label{eq:openaiesgrad}
\end{align}
\openaies{} then passes this estimate to an Adam optimizer \citep{adam} which outputs a gradient ascent step for $\vphi$. To make the estimate more accurate, \openaies{} further includes techniques such as mirror sampling and rank normalization \citep{brockhoff2010mirror,ha2017visual,wierstra2014nes}.

\subsubsection{Actor-critic methods} While ES treats the objective as a black box, actor-critic methods leverage the MDP structure of the objective, i.e. the fact that $f(\vphi)$ is a sum of Markovian values. We are most interested in Twin Delayed Deep Deterministic policy gradient (TD3) \citep{fujimoto2018td3}, an off-policy actor-critic method. TD3 maintains (1) an actor consisting of the policy $\pi_\vphi$ and (2) a critic consisting of state-action value functions $Q_{\vtheta_1}(s, a)$ and $Q_{\vtheta_2}(s, a)$ which differ only in random initialization. Through interactions in the environment, the actor generates experience which is stored in a replay buffer $\mathcal{B}$. This experience is sampled to train $Q_{\vtheta_1}$ and $Q_{\vtheta_2}$. Simultaneously, the actor improves by maximizing $Q_{\vtheta_1}$ via gradient ascent ($Q_{\vtheta_2}$ is only used during critic training). Specifically, for an objective $f'$ which is based on the critic and approximates $f$, TD3 estimates a gradient $\vnabla f'(\vphi)$ and passes it to an Adam optimizer. Notably, TD3 never updates network weights directly, instead accumulating weights into \textit{target networks} $\pi_{\vphi'}$, $Q_{\vtheta'_1}$, $Q_{\vtheta'_2}$ via an exponentially weighted moving average with update rate $\tau$.

\subsection{Quality Diversity Algorithms} \label{sec:qd}

\subsubsection{\mapelites{} extensions for \qdrl{}} \label{sec:mapelites} One of the simplest QD algorithms is \mapelites{} \citep{mouret2015illuminating, cully2015}. \mapelites{} creates an archive $\mathcal{A}$ by tesselating the measure space $\mathcal{X}$ into a grid of evenly-sized cells. Then, it draws $\lambda$ initial solutions from a multivariate Gaussian $\mathcal{N}(\mathbf{\vphi_0}, \sigma \mI)$ centered at some $\vphi_0$. Next, for each sampled solution $\vphi$, \mapelites{} computes $f(\vphi)$ and $\vm(\vphi)$ and inserts $\vphi$ into $\mathcal{A}$. In subsequent iterations, \mapelites{} randomly selects $\lambda$ solutions from $\mathcal{A}$ and adds Gaussian noise, i.e. solution $\vphi$ becomes $\vphi + \mathcal{N}(\mathbf{0}, \sigma \mI)$. Solutions are placed into cells based on their measures; if a solution has higher $f$ than the solution currently in the cell, it replaces that solution. Once inserted into $\mathcal{A}$, solutions are known as \textit{elites}.

Due to the high dimensionality of neural network parameters, direct policy optimization with \mapelites{} has not proven effective in \qdrl{} \citep{colas2020scaling}, although indirect encodings have been shown to scale to large policy networks \citep{rakicevic2021poms,gaier2020blackbox}. For direct search, several extensions merge \mapelites{} with actor-critic methods and ES. For instance, Policy Gradient Assisted \mapelites{} (\pgame{}) \citep{nilsson2021pga} combines \mapelites{} with TD3. Each iteration, \pgame{} evaluates $\lambda$ solutions for insertion into the archive. $\frac{\lambda}{2}$ of these are created by selecting random solutions from the archive and taking gradient ascent steps with a TD3 critic. The other $\frac{\lambda}{2}$ solutions are created with a directional variation operator \citep{vassiliades2018line} which selects two solutions $\vphi_1$ and $\vphi_2$ from the archive and creates a new one according to $\vphi' = \vphi_1 + \sigma_1 \mathcal{N}(\mathbf{0}, \mI) + \sigma_2 (\vphi_2 - \vphi_1) \mathcal{N}(0, 1)$. Finally, \pgame{} maintains a ``greedy actor'' which provides actions when training the critics (identically to the actor in TD3). Every iteration, \pgame{} inserts this greedy actor into the archive. \pgame{} achieves state-of-the-art performance on locomotion tasks in the QDGym benchmark \citep{qdgym}.

Another \mapelites{} extension is \mees{} \citep{colas2020scaling}, which combines \mapelites{} with an \openaies{} optimizer. In the ``explore-exploit'' variant, \mees{} alternates between two phases. In the ``exploit'' phase, \mees{} restarts \openaies{} at a mean $\vphi$ and optimizes the objective for $k$ iterations, inserting the current $\vphi$ into the archive in each iteration. In the ``explore'' phase, \mees{} repeats this process, but \openaies{} instead optimizes for novelty, where novelty is the distance in measure space from a new solution to previously encountered solutions. \mees{} also has an ``exploit'' variant and an ``explore'' variant, which each execute only one type of phase.

Our work is related to \mees{} in that we also adapt \openaies{}, but instead of alternating between following a novelty gradient and objective gradient, we compute all objective and measure gradients and allow a \cmaes{} \citep{hansen2016tutorial} instance to decide which gradients to follow by sampling gradient coefficients from a multivariate Gaussian updated over time (\sref{sec:cmamega}). We include \mapelites{}, \pgame{}, and \mees{} as baselines in our experiments. Refer to \fref{fig:tree} for a diagram which compares these algorithms to our approach.

\subsubsection{Covariance Matrix Adaptation MAP-Elites via a Gradient Arborescence (\cmamega{})} \label{sec:cmamega} We directly extend \cmamega{} \citep{fontaine2021dqd} to address \qdrl{}. \cmamega{} is a DQD algorithm based on the QD algorithm \cmame{} \citep{fontaine2020covariance}. The intuition behind \cmamega{} is that if we knew which direction the current solution point $\vphi^*$ should move in objective-measure space, then we could calculate that change in search space via a linear combination of objective and measure gradients. From \cmame{}, we know a good direction is one that results in the largest archive improvement.

Each iteration, \cmamega{} first calculates objective and measure gradients for a solution point $\vphi^*$. Next, it generates $\lambda$ new solutions by sampling gradient coefficients $\vc \sim \mathcal{N}(\vmu, \mSigma)$ and computing $\vphi' \gets \vphi^* + c_0 \vnabla f(\vphi^*) + \sum_{j=1}^k c_j \vnabla m_j (\vphi^*)$. \cmamega{} inserts these solutions into the archive and computes their \textit{improvement,} $\Delta$. $\Delta$ is defined as $f(\vphi')$ if $\vphi'$ populates a new cell, and $f(\vphi') - f(\vphi'_\mathcal{E})$ if $\vphi'$ improves an existing cell (replaces a previous solution $\vphi'_\mathcal{E}$). After \cmamega{} inserts the solutions, it ranks them by $\Delta$. If a solution populates a new cell, its $\Delta$ always ranks higher than that of a solution which only improves an existing cell. \cmamega{} then moves the solution point $\vphi^*$ towards the largest archive improvement, but also adapts the distribution $\mathcal{N}(\vmu, \mSigma)$ towards better gradient coefficients by the same ranking. By leveraging gradient information, \cmamega{} solves QD benchmarks with orders of magnitude fewer solution evaluations than previous QD algorithms.

\subsubsection{Beyond \mapelites{}} Several \qdrl{} algorithms have been developed outside the \mapelites{} family. \nses{} \citep{conti2018ns} builds on Novelty Search (NS) \citep{lehman2011ns, lehman2011nslc}, a family of QD algorithms which add solutions to an \textit{unstructured archive} only if they are far away from existing archive solutions in measure space. Using \openaies{}, \nses{} concurrently optimizes several agents for novelty. Its variants \nsres{} and \nsraes{} optimize for a linear combination of novelty and objective. Meanwhile, the \qdrl{} algorithm \citep{cideron2020qdrl} (distinct from the \qdrl{} problem we define) maintains an archive with all past solutions and optimizes agents along a Pareto front of the objective and novelty. Finally, Diversity via Determinants (DvD) \citep{parkerholder2020dvd} leverages a kernel method to maintain diversity in a population of solutions. As \nses{}, \qdrl{}, and DvD do not output a \mapelites{} grid archive, we leave their investigation for future work.

\subsection{Diversity in Reinforcement Learning}

Here we distinguish \qdrl{} from prior work which also applies diversity to RL. One area of work is in latent- and goal-conditioned policies. For latent-conditioned policy $\pi_\vphi(a | s, z)$ \citep{eysenbach2019diversity,kumar2020one,li2017infogail} or goal-conditioned policy $\pi_\vphi(a | s, g)$ \citep{schaul2015uvfa,andrychowicz2017her}, varying the latent variable $z$ or goal $g$ results in different behaviors, e.g. different walking gaits or walking to a different location. While \qdrl{} also seeks a range of behaviors, the measures $\vm(\vphi)$ are computed \textit{after} evaluating $\vphi$, rather than \textit{before} the evaluation. In general, \qdrl{} focuses on finding a variety of policies for a single task, rather than attempting to solve a variety of tasks with a single conditioned policy.

Another area of work combines evolutionary and actor-critic algorithms to solve single-objective hard-exploration problems \citep{colas2018geppg,khadka2018erl,pourchot2018cemrl,tang2021cemacer,khadka2019cerl}. In these methods, an evolutionary algorithm such as cross-entropy method \citep{cem} facilitates exploration by generating a diverse population of policies, while an actor-critic algorithm such as TD3 trains high-performing policies with this population's environment experience. \qdrl{} differs from these methods in that it views diversity as a component of the output, while these methods view diversity as a means for environment exploration. Hence, \qdrl{} measures policy behavior via a measure function and collects diverse policies in an archive. In contrast, these RL exploration methods assume that trajectory diversity, rather than targeting specific behavioral diversity, is enough to drive exploration to discover a single optimal policy.

\begin{figure}[t]
\ifModeWebsite{\lxAddClass{mount_tree}}
\begin{center}
  \ifdefined\modeJournal
    \includegraphics[width=0.8\linewidth]{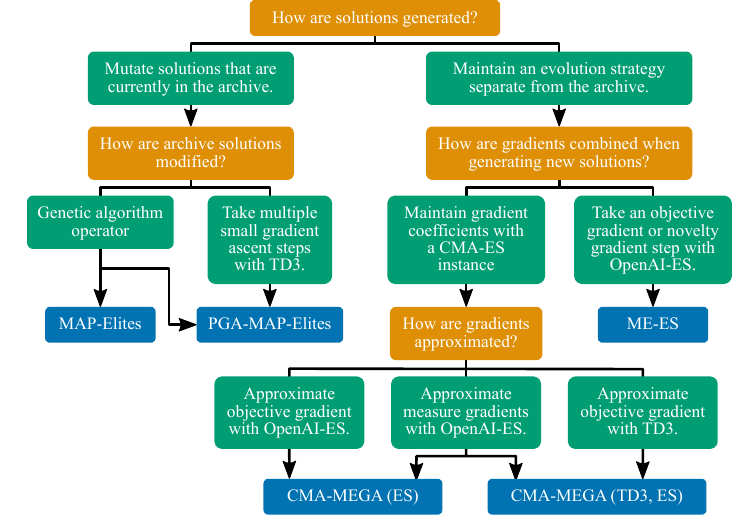}
  \else
    \ifdefined\modeIclr
      \includegraphics[width=0.5\linewidth]{imgs/tree/tree.pdf}
    \else
      \includegraphics[width=\linewidth]{imgs/tree/tree.pdf}
    \fi
  \fi
\end{center}
\caption{Diagram of \mapelites{} extensions for \qdrl{}, showing how our \cmamega{} variants differ from other \qdrl{} algorithms.}
\label{fig:tree}
\Description[Diagram of \mapelites{} extensions for \qdrl{}.]{This diagram is a tree diagram with questions and answers that lead to each algorithm. At the top of the tree is the question "How are solutions generated?" On the left is the answer "Mutate solutions that are currently in the archive." In this case, the next question is "How are archive solutions modified?" If the answer is "Genetic algorithm operator," we arrive at \mapelites{}. If the answer is "Take multiple small gradient ascent steps with TD3," we arrive at \pgame{}. Note that \pgame{} also uses a genetic algorithm operator for some of its solutions. Going back to the original question ("How are solutions generated?"), the answer on the right says "Maintain an evolution strategy separate from the archive." In this case, the next question is "How are gradients combined when generating new solutions?" One answer is "Take an objective gradient or novelty gradient step with OpenAI-ES," in which case we arrive at the ME-ES algorithm. The other answer is "Maintain gradient coefficients with a CMA-ES instance," which leads to the question "How are gradients approximated?" Here, if we "Approximate objective gradient with OpenAI-ES" and "Approximate measure gradients with OpenAI-ES," we arrive at the CMA-MEGA (ES) algorithm. If we instead "Approximate objective gradient with TD3" and "Approximate measure gradients with OpenAI-ES," we arrive at the CMA-MEGA (TD3, ES) algorithm.}
\end{figure}

\section{Approximating Gradients for DQD} \label{sec:cmamegaapprox}

Since DQD algorithms require exact objective and measure gradients, we cannot directly apply \cmamega{} to \qdrl{}. To address this limitation, we replace exact gradients with gradient approximations (\sref{sec:approx}) and develop two \cmamega{} variants (\sref{sec:cmamegavariants}).

\subsection{Approximating Objective and Measure Gradients} \label{sec:approx}

We adapt gradient approximations from ES and actor-critic methods. Since the objective has an MDP structure, we estimate objective gradients $\vnabla f$ with ES and actor-critic methods. Since the measures are black boxes, we estimate measure gradients $\vnabla \vm$ with ES.

\subsubsection{Approximating objective gradients with ES and actor-critic methods}

We estimate objective gradients with two methods. First, we treat the objective as a black box and estimate its gradient with a black box method, namely the \openaies{} gradient estimate in \eref{eq:openaiesgrad}. Since \openaies{} performs well in RL domains \citep{salimans2017evolution,pagliuca2020,lehman2018}, we believe this estimate is suitable for approximating gradients for \cmamega{} in \qdrl{} settings. Importantly, this estimate requires environment interaction by evaluating $\lambda_{es}$ solutions.

Since the objective has a well-defined structure, i.e. it is a sum of rewards from an MDP (\eref{eq:return}), we also estimate its gradient with an actor-critic method, TD3. TD3 is well-suited for this purpose because it efficiently estimates objective gradients for the multiple policies that \cmamega{} and other \qdrl{} algorithms generate. In particular, once the critic is trained, TD3 can provide a gradient estimate for any policy without additional environment interaction.

Among actor-critic methods, we select TD3 since it achieves high performance while optimizing primarily for the RL objective. Prior work \citep{fujimoto2018td3} shows that TD3 outperforms on-policy actor-critic methods \citep{schulman2015trpo,schulman2017ppo}. While the off-policy Soft Actor-Critic \citep{haarnoja2018sac} algorithm can outperform TD3, it optimizes a maximum-entropy objective designed to encourage exploration. %
In our work, we explore by finding policies with different measures. Thus, we leave for future work the problem of integrating \qdrl{} with the action diversity encouraged by entropy maximization.

\subsubsection{Approximating measure gradients with ES} \label{sec:measuregrad}

Since measures do not have special properties such as an MDP structure (\sref{sec:qdrl}), we only estimate their gradient with black box methods. Thus, similar to the objective, we approximate each measure's gradient $\vnabla {m_i}$ with the \openaies{} gradient estimate, replacing $f$ with $m_i$ in \eref{eq:openaiesgrad}.

Since the \openaies{} gradient estimate requires additional environment interaction, all of our \cmamega{} variants require environment interaction to estimate gradients. However, the environment interaction required to estimate measure gradients remains constant even as the number of measures increases, since we can reuse the same $\lambda_{es}$ solutions to estimate each $\vnabla {m_i}$.

In problems where the measures have an MDP structure similar to the objective, it may be feasible to estimate each $\vnabla {m_i}$ with its own TD3 instance. In the environments in our work (\sref{sec:domains}), each measure is non-Markovian since it calculates the proportion of time a walking agent's foot spends on the ground. This calculation depends on the entire agent trajectory rather than on one state.

\subsection{\cmamega{} Variants} \label{sec:cmamegavariants}

\ifModeWebsite{\begin{algorithm}[t]

\ifdefined\modeWebsite
  \caption{\cmamegaes{} and \cmamegatdes{}. \hlc[lightgrey]{Highlighted portions} are only executed in \cmamegatdes{}. Adapted from \cmamega{}. Refer to the appendix for methods whose names are in \textsc{Small\_Caps}.}
\else
  \caption{\cmamegaes{} and \cmamegatdes{}. \hlc[lightgrey]{Highlighted portions} are only executed in \cmamegatdes{}. Adapted from \cmamega{} \citep{fontaine2021dqd}. Refer to \apref{sec:extra-code} for functions whose names are in \textsc{Small\_Caps}.}
\fi

\label{alg:cmamegaes}

\SetKwProg{CmaMega}{CMA-MEGA variants}{:}{}
\CmaMega{$(evaluate, \vphi_0, N, \lambda, \sigma_g, \eta, \lambda_{es}, \sigma_e)$}{
  \KwIn{Function $evaluate$ which executes a policy $\vphi$ and outputs objective $f(\vphi)$ and measures $\vm(\vphi)$, initial solution $\vphi_0$, desired iterations $N$, batch size $\lambda$, initial \cmaes{} step size $\sigma_g$, learning rate $\eta$, ES batch size $\lambda_{es}$, ES standard deviation $\sigma_e$}
  \KwResult{Generates $N\lambda$ solutions, storing elites in an archive $\mathcal{A}$}

  $\lambda' \gets \lambda - 1 $\hlc[lightgrey]{$\ -\ 1$}\; \label{alg:cmamegaes:reduce}

  Initialize empty archive $\mathcal{A}$, solution point $\vphi^* \gets \vphi_0$ \label{alg:cmamegaes:init} \;

  Initialize \cmaes{} with population $\lambda'$, resulting in $\vmu = \mathbf{0}, \mSigma = \sigma_g \mI$, and internal \cmaes{} parameters $\vp$ \label{alg:cmamegaes:cmaesinit} \;

  \hlc[lightgrey]{$\mathcal{B}, Q_{\vtheta_1}, Q_{\vtheta_2}, \pi_{\vphi_q}, Q_{\vtheta'_1}, Q_{\vtheta'_2}, \pi_{\vphi'_q} \gets$ \textsc{Initialize\_TD3()}} \label{alg:cmamegaes:inittd} \;

  \For{$iter \gets 1..N$}{ \label{alg:cmamegaes:loop}
    $f(\vphi^*), \vm(\vphi^*) \gets evaluate(\vphi^*)$ \; \label{alg:cmamegaes:eval-mean}
    \textsc{Update\_Archive}($\mathcal{A}, \vphi^*, f(\vphi^*), \vm(\vphi^*)$) \; \label{alg:cmamegaes:insert-mean}
    $\vnabla f(\vphi^*), \vnabla \vm(\vphi^*) \gets$ \textsc{ES\_Gradients}($\vphi^*, \lambda_{es}, \sigma_e$) \; \label{alg:cmamegaes:es-estimate}
    \hlc[lightgrey]{$\vnabla f(\vphi^*) \gets$ \textsc{TD3\_Gradient($\vphi^*, Q_{\vtheta_1}, \mathcal{B}$)}} \; \label{alg:cmamegaes:td3-estimate}
    Normalize $\vnabla f(\vphi^*)$ and $\vnabla \vm(\vphi^*)$ to be unit vectors \label{alg:cmamegaes:norm}

    \For{$i \gets 1..\lambda'$} { \label{alg:cmamegaes:pop-loop}
      $\vc_i \sim \mathcal{N} (\vmu, \mSigma)$ \; \label{alg:cmamegaes:coeff}
      $\vnabla_i \gets c_{i,0} \vnabla f(\vphi^*) + \sum_{j=1}^k c_{i,j} \vnabla m_j(\vphi^*)$ \; \label{alg:cmamegaes:make-nabla}
      $\vphi'_i \gets \vphi^* + \vnabla_i$ \; \label{alg:cmamegaes:apply-nabla}
      $f(\vphi'_i), \vm'(\vphi'_i) \gets evaluate(\vphi'_i)$ \; \label{alg:cmamegaes:eval-phi}
      $\Delta_i \gets$ \textsc{Update\_Archive}($\mathcal{A}, \vphi'_i, f(\vphi'_i), \vm(\vphi'_i)$) \; \label{alg:cmamegaes:insert-phi}
    }

    Rank $\vc_i, \vnabla_i$ by $\Delta_i$ \; \label{alg:cmamegaes:rank}
    Adapt CMA-ES parameters $\vmu, \mSigma, \vp$ based on rankings of $\vc_i$ \; \label{alg:cmamegaes:adapt}

    $\vphi^* \gets \vphi^* + \eta \sum_{i=1}^\lambda w_i \vnabla_{\text{rank[i]}}$ \tcp{$w_i$ is part of $\vp$} \label{alg:cmamegaes:step}

    \If{there is no change in $\mathcal{A}$} { \label{alg:cmamegaes:restart}
      Restart CMA-ES with $\vmu=\mathbf{0},\mSigma=\sigma_g\mI$ \; \label{alg:cmamegaes:restart-cmaes}
      Set $\vphi^*$ to a randomly selected elite from $\mathcal{A}$ \; \label{alg:cmamegaes:select}
    }

    \hlc[lightgrey]{$f(\vphi_q),\vm(\vphi_q) \gets evaluate(\vphi_q)$} \; \label{alg:cmamegaes:greedy-eval}
    \hlc[lightgrey]{\textsc{Update\_Archive}($\mathcal{A}, \vphi_q, f(\vphi_q), \vm(\vphi_q)$)} \; \label{alg:cmamegaes:greedy-insert}
    \hlc[lightgrey]{Add experience from all calls to $evaluate$ into $\mathcal{B}$} \; \label{alg:cmamegaes:buffer}
    \hlc[lightgrey]{\textsc{Train\_TD3($Q_{\vtheta_1}, Q_{\vtheta_2}, \pi_{\vphi_q}, Q_{\vtheta'_1}, Q_{\vtheta'_2}, \pi_{\vphi'_q}, \mathcal{B}$)}} \; \label{alg:cmamegaes:train}
  }
}
\end{algorithm}

}
\ifModeQuals{}
\ifModeQualsOff{\ifModeWebsiteOff{}}

Our choice of gradient approximations leads to two \cmamega{} variants. \textbf{\cmamegaes{}} approximates objective and measure gradients with \openaies{}, while \textbf{\cmamegatdes{}} approximates the objective gradient with TD3 and measure gradients with \openaies{}. \fref{fig:diagram} shows an overview of both algorithms, and \aref{alg:cmamegaes} shows their pseudocode. As \cmamegatdes{} builds on \cmamegaes{}, we present only \cmamegatdes{}  and highlight lines that \cmamegatdes{} additionally executes.

\newcommand{\lineRef}[2]{\ifdefined\modeWebsite#2\else\ref{#1}\fi}

Identically to \cmamega{}, the two variants maintain three primary components: a solution point $\vphi^*$, a multivariate Gaussian distribution $\mathcal{N}(\vmu, \mSigma)$ for sampling gradient coefficients, and a \mapelites{} archive $\mathcal{A}$ for storing solutions. We initialize the archive and solution point on line \lineRef{alg:cmamegaes:init}{3}, and we initialize the coefficient distribution as part of a \cmaes{} instance on line \lineRef{alg:cmamegaes:cmaesinit}{4}.\footnote{We set the \cmaes{} batch size $\lambda'$ slightly lower than the total batch size $\lambda$ (line \lineRef{alg:cmamegaes:reduce}{2}). While \cmamegaes{} and \cmamegatdes{} both evaluate $\lambda$ solutions each iteration, one evaluation is reserved for $\vphi^*$ (line \lineRef{alg:cmamegaes:eval-mean}{7}). In \cmamegatdes{}, one more evaluation is reserved for the greedy actor (line \lineRef{alg:cmamegaes:greedy-eval}{26}).}

In the main loop (line \lineRef{alg:cmamegaes:loop}{6}), we follow the workflow shown in \fref{fig:diagram}. First, after evaluating $\vphi^*$ and inserting it into the archive (line \lineRef{alg:cmamegaes:eval-mean}{7}-\lineRef{alg:cmamegaes:insert-mean}{8}), we approximate its gradients with either ES or TD3 (line \lineRef{alg:cmamegaes:es-estimate}{9}-\lineRef{alg:cmamegaes:td3-estimate}{10}). \textit{This gradient approximation forms the key difference between our variants and the original \cmamega{} algorithm \citep{fontaine2021dqd}.}

Next, we branch from $\vphi^*$ to create solutions $\vphi'_i$ by sampling $\vc_i$ from the coefficient distribution and computing perturbations $\vnabla_i$ (line \lineRef{alg:cmamegaes:coeff}{13}-\lineRef{alg:cmamegaes:apply-nabla}{15}). We then evaluate each $\vphi'_i$ and insert it into the archive (line \lineRef{alg:cmamegaes:eval-phi}{16}-\lineRef{alg:cmamegaes:insert-phi}{17}).

Finally, we update the solution point and the coefficient distribution's \cmaes{} instance by forming an \textit{improvement ranking} based on the improvement $\Delta_i$ (\sref{sec:cmamega}; line \lineRef{alg:cmamegaes:rank}{19}-\lineRef{alg:cmamegaes:step}{21}). Importantly, since we rank based on improvement, this update enables the \cmamega{} variants to maximize the QD objective (\eref{eq:qdobjective}) \citep{fontaine2021dqd}.

Our \cmamega{} variants have two additional components. First, we check if no solutions were inserted into the archive at the end of the iteration, which would indicate that we should reset the coefficient distribution and the solution point (line \lineRef{alg:cmamegaes:restart}{22}-\lineRef{alg:cmamegaes:select}{24}). Second, in the case of \cmamegatdes{}, we manage a TD3 instance similar to how \pgame{} does (\sref{sec:mapelites}). This TD3 instance consists of a replay buffer $\mathcal{B}$, critic networks $Q_{\vtheta_1}$ and $Q_{\vtheta_2}$, a greedy actor $\pi_{\vphi_q}$, and target networks $Q_{\vtheta'_1}$, $Q_{\vtheta'_2}$, $\pi_{\vphi'_q}$ (all initialized on line \lineRef{alg:cmamegaes:inittd}{5}). At the end of each iteration, we use the greedy actor to train the critics, and we also insert it into the archive (line \lineRef{alg:cmamegaes:greedy-eval}{26}-\lineRef{alg:cmamegaes:train}{29}).

\section{Experiments} \label{sec:experiments}

We compare our two proposed \cmamega{} variants (\cmamegaes{}, \cmamegatdes{}) with three baselines (\pgame{}, \mees{}, \mapelites{}) in four locomotion tasks. We implement \mapelites{} as described in \sref{sec:mapelites}, and we select the explore-exploit variant for \mees{} since it has performed at least as well as both the explore variant and the exploit variant in several domains \citep{colas2020scaling}.

\subsection{Evaluation Domains} \label{sec:domains}

\subsubsection{QDGym} \label{sec:qdgym}

We evaluate our algorithms in four locomotion environments from QDGym \citep{qdgym}, a library built on PyBullet Gym \citep{coumans2020, benelot2018} and OpenAI Gym \citep{brockman2016gym}. %
\apref{sec:environment_details} lists all environment details. In each environment, the QD algorithm outputs an archive of walking policies for a simulated agent. The agent is primarily rewarded for its forward speed. There are also reward shaping \citep{ng1999shaping} signals, such as a punishment for applying higher joint torques, intended to guide policy optimization. The measures compute the proportion of time (number of timesteps divided by total timesteps in an episode) that each of the agent's feet contacts the ground.

QDGym is challenging because the objective in each environment does not ``align'' with the measures, in that finding policies with different measures (i.e. exploring the archive) does not necessarily lead to optimization of the objective. While it may be trivial to fill the archive with low-performing policies which stand in place and lift the feet up and down to achieve different measures, the agents' complexity (high degrees of freedom) makes it difficult to learn a high-performing policy for each value of the measures.

\ifModeWebsite{\begin{figure}[t]

\begin{center}
\begin{tabular}{lrrrr}

{} &
\multicolumn{1}{p{0.12\linewidth}}{\centering QD Ant}
   &
\multicolumn{1}{p{0.3\linewidth}}{\centering QD Half-Cheetah}
   &
\multicolumn{1}{p{0.18\linewidth}}{\centering QD Hopper}
   &
\multicolumn{1}{p{0.18\linewidth}}{\centering QD Walker}
\vspace{4pt}
\\

{} &
\multicolumn{1}{c}{\includegraphics[width=0.12\linewidth]{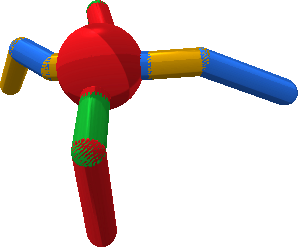}}
   &
\multicolumn{1}{c}{\includegraphics[width=0.12\linewidth]{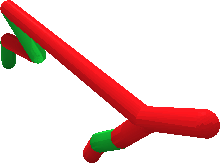}}
   &
\multicolumn{1}{c}{\includegraphics[width=0.036\linewidth]{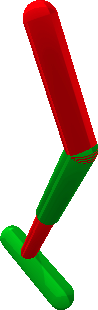}}
   &
\multicolumn{1}{c}{\includegraphics[width=0.05\linewidth]{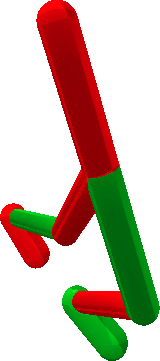}}
\\
\end{tabular}
\end{center}
\caption{QDGym locomotion environments \citep{qdgym}.}
\label{fig:envs}

\end{figure}

}

\subsubsection{Hyperparameters} \label{sec:envparams}

Each agent's policy is a neural network which takes in states and outputs actions. There are two hidden layers of 128 nodes, and the hidden and output layers have \texttt{tanh} activation. We initialize weights with Xavier initialization \citep{glorot2010}.

For the archive, we tesselate each environment's measure space into a grid of evenly-sized cells (see \tref{table:envs} for grid dimensions). Each measure is bound to the range $[0, 1]$, the min and max proportion of time that one foot can contact the ground.

Each algorithm evaluates 1 million solutions in the environment. Due to computational limits, we evaluate each solution once instead of averaging multiple episodes, so each algorithm runs 1 million episodes total. Refer to \apref{sec:algoparams} for further hyperparameters.

\subsubsection{Metrics} \label{sec:metrics}

Our primary metric is \textit{QD score} \citep{pugh2016qd}, which provides a holistic view of algorithm performance. QD score is the sum of the objective values of all elites in the archive, i.e. $\sum_{i=1}^M \1_{\vphi_i \mathrm{exists}} f(\vphi_i)$, where $M$ is the number of archive cells. We note that the contribution of a cell to the QD score is 0 if the cell is unoccupied. We set the objective $f$ to be the \textit{expected undiscounted return}, i.e. we set $\gamma = 1$ in \eref{eq:return}.

Since objectives may be negative, an algorithm's QD score may be penalized when adding a new solution. To prevent this, we define a \textit{minimum objective} in each environment by taking the lowest objective value that was inserted into the archive in any experiment in that environment. We subtract this minimum from every solution, such that every solution that was inserted into an archive has an objective value of at least 0. Thus, we use QD score defined as $\sum_{i=1}^M \1_{\vphi_i \mathrm{exists}} (f(\vphi_i) - \mathrm{min\ objective})$.
We also define a \textit{maximum objective} equivalent to each environment's ``reward threshold'' in PyBullet Gym. This threshold is the objective value at which an agent is considered to have successfully learned to walk.

We report two metrics in addition to QD score. \textit{Archive coverage}, the proportion of cells for which the algorithm found an elite, gauges how well the QD algorithm explores measure space, and \textit{best performance}, the highest objective of any elite in the archive, gauges how well the QD algorithm exploits the objective.

\ifModeWebsiteOff{\begin{figure}[t]
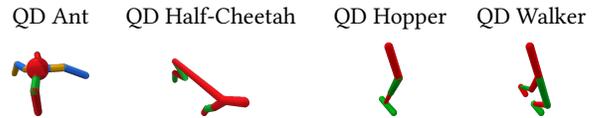


\begin{center}
\begin{tabular}{lrrrr}

{} &
\multicolumn{1}{p{0.12\linewidth}}{\centering QD Ant}
   &
\multicolumn{1}{p{0.3\linewidth}}{\centering QD Half-Cheetah}
   &
\multicolumn{1}{p{0.18\linewidth}}{\centering QD Hopper}
   &
\multicolumn{1}{p{0.18\linewidth}}{\centering QD Walker}
\vspace{4pt}
\\

{} &
\multicolumn{1}{c}{\includegraphics[width=0.12\linewidth]{imgs/qdgym_env_imgs/ant.png}}
   &
\multicolumn{1}{c}{\includegraphics[width=0.12\linewidth]{imgs/qdgym_env_imgs/half_cheetah.png}}
   &
\multicolumn{1}{c}{\includegraphics[width=0.036\linewidth]{imgs/qdgym_env_imgs/hopper.png}}
   &
\multicolumn{1}{c}{\includegraphics[width=0.05\linewidth]{imgs/qdgym_env_imgs/walker.png}}
\\
\end{tabular}
\end{center}
\caption{QDGym locomotion environments \citep{qdgym}.}
\label{fig:envs}

\end{figure}

}

\subsection{Experimental Design} \label{sec:experimentdesign}

We follow a between-groups design, where the two independent variables are environment (QD Ant, QD Half-Cheetah, QD Hopper, QD Walker) and algorithm (\cmamegaes{}, \cmamegatdes{}, \pgame{}, \mees{}, \mapelites{}). The dependent variable is the QD score. In each environment, we run each algorithm for 5 trials with different random seeds and test three hypotheses:

\textbf{H1:} \cmamegaes{} will outperform  all baselines (\pgame{}, \mees{}, \mapelites{}).

\textbf{H2:} \cmamegatdes{} will outperform all baselines.

\textbf{H3:}  \cmamegatdes{} will outperform \cmamegaes{}.

H1 and H2 are based on prior work \citep{fontaine2021dqd} which showed that in QD benchmark domains, \cmamega{} outperforms algorithms that do not leverage both objective and measure gradients. H3 is based on results \citep{pagliuca2020} which suggest that actor-critic methods outperform ES in PyBullet Gym.
Thus, we expect the TD3 objective gradient to be more accurate than the ES objective gradient, leading to more efficient traversal of objective-measure space and higher QD score.

\subsection{Implementation} \label{sec:impl}

We implement all QD algorithms with the pyribs library \citep{pyribs} except for \mees{}, which we adapt from the authors' implementation. We run each experiment with 100 CPUs on a high-performance cluster. We allocate one NVIDIA Tesla P100 GPU to algorithms that train TD3 (\cmamegatdes{} and \pgame{}). Depending on the algorithm and environment, each experiment lasts 4-20 hours; refer to \tref{table:runtime-hours}, \apref{sec:finalmetrics} for mean runtimes.
\begin{anonsuppress}
We have released our source code at \url{https://github.com/icaros-usc/dqd-rl} %
\end{anonsuppress}

\section{Results} \label{sec:results}

\begin{figure*}[t]
\ifModeWebsite{\lxAddClass{mount_comparison}}
\begin{center}
\includegraphics[width=\textwidth]{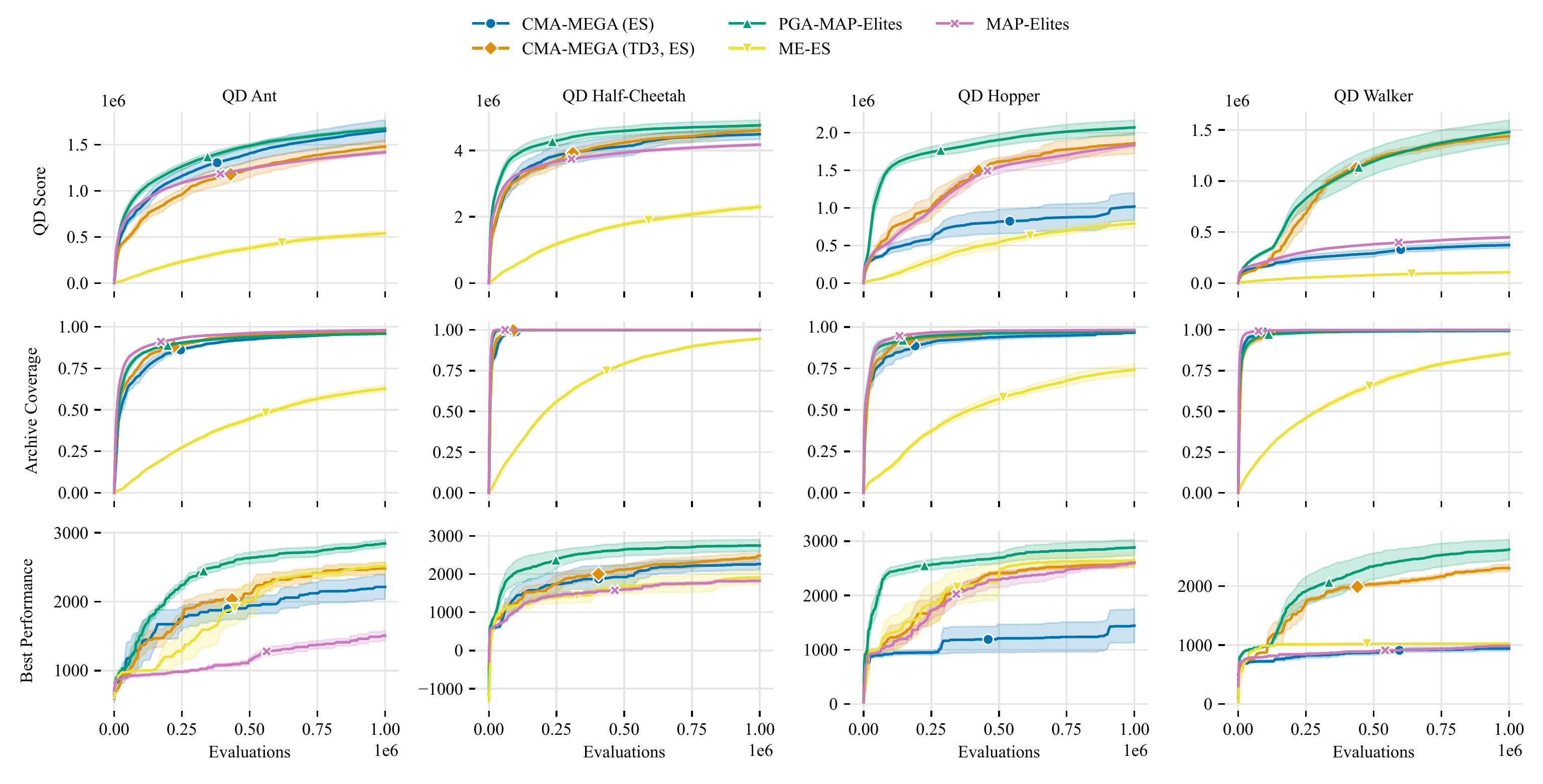}
\end{center}
\caption{Plots of QD score, archive coverage, and best performance for the 5 algorithms in our experiments in all 4 environments from QDGym. The x-axis in all plots is the number of solutions evaluated. Solid lines show the mean over 5 trials, and shaded regions show the standard error of the mean.}
\label{fig:comparison}
\Description[Fully described in the text]{}
\end{figure*}

We ran 5 trials of each algorithm in each environment. In each trial, we allocated 1 million evaluations and recorded the QD score, archive coverage, and best performance. \fref{fig:comparison} plots these metrics, and \apref{sec:finalmetrics} lists final values of all metrics. \apref{sec:archives} shows example heatmaps and histograms of each archive, and the supplemental material contains videos of generated agents.

\subsection{Analysis} \label{sec:analysis}

To test our hypotheses, we conducted a two-way ANOVA which examined the effect of algorithm and environment on the QD score. We note that the ANOVA requires QD scores to have the same scale, but each environment's QD score has a different scale by default. Thus, for this analysis, we normalized QD scores by dividing by each environment's maximum QD score, defined as \textit{grid cells} * (\textit{max objective} - \textit{min objective}) (see \apref{sec:environment_details} for these quantities).

We found a statistically significant interaction between algorithm and environment on QD score, $F(12,80) = 16.82, p < 0.001$. Simple main effects analysis indicated that the algorithm had a significant effect on QD score in each environment, so we ran pairwise comparisons (two-sided t-tests) with Bonferroni corrections (\apref{sec:fullstats}). Our results are as follows:

\textbf{H1:} There is no significant difference in QD score between \cmamegaes{} and \pgame{} in QD Ant and QD Half-Cheetah, but in QD Hopper and QD Walker, \cmamegaes{} attains significantly lower QD score than \pgame{}. \cmamegaes{} achieves significantly higher QD score than \mees{} in all environments except QD Hopper, where there is no significant difference. There is no significant difference between \cmamegaes{} and \mapelites{} in all domains except QD Hopper, where \cmamegaes{} attains significantly lower QD score.

\textbf{H2:} In all environments, there is no significant difference in QD score between \cmamegatdes{} and \pgame{}. \cmamegatdes{} achieves significantly higher QD score than \mees{} in all environments. \cmamegatdes{} achieves significantly higher QD score than \mapelites{} in QD Half-Cheetah and Walker, with no significant difference in QD Ant and QD Hopper.

\textbf{H3:} \cmamegatdes{} achieves significantly higher QD score than \cmamegaes{} in QD Hopper and QD Walker, but there is no significant difference in QD Ant and QD Half-Cheetah.

\subsection{Discussion}

We discuss how the \cmamega{} variants differ from the baselines (\sref{sec:results-pgame}-\ref{sec:results-mapelites}) and how they differ from each other (\sref{sec:results-variants}).

\subsubsection{\pgame{} and objective-measure space exploration} \label{sec:results-pgame}

Of the \cmamega{} variants, \cmamegatdes{} performed the closest to \pgame{}, with no significant QD score difference in any environment. This result differs from prior work \citep{fontaine2021dqd} in QD benchmark domains, where \cmamega{} outperformed \ogmapelites{}, a baseline DQD algorithm inspired by \pgame{}.

We attribute this difference to the difficulty of exploring objective-measure space in the benchmark domains. For example, the linear projection benchmark domain is designed to be ``distorted'' \citep{fontaine2020covariance}. Values in the center of its measure space are easy to obtain with random sampling, while values at the edges are unlikely to be sampled. Hence, high QD score arises from exploring measure space and filling the archive. Since \cmamega{} adapts its sampling distribution, it is able to perform this exploration, while \ogmapelites{} remains ``stuck'' in the center of the measure space.

In contrast, as discussed in \sref{sec:qdgym}, it is relatively easy to fill the archive in QDGym. We see this empirically: in all environments, all algorithms achieve nearly 100\% archive coverage, usually within the first 250k evaluations (\fref{fig:comparison}). Hence, the best QD score is achieved by increasing the objective value of solutions after filling the archive.  \pgame{} achieves this by optimizing half of its generated solutions with respect to its TD3 critic. The genetic operator likely further enhances the efficacy of this optimization, by taking previously-optimized solutions and combining them to obtain high-performing solutions in other parts of the archive.

On the other hand, the \cmamega{} variants place less emphasis on maximizing the performance of each solution, compared to \pgame{}: in each trial, \pgame{} takes 5 million objective gradient steps with respect to its TD3 critic, while the \cmamega{} variants only compute 5k objective gradients, because they dedicate a large part of the evaluation to estimating the measure gradients. This difference suggests a possible extension to \cmamegatdes{} in which solutions are optimized with respect to the TD3 critic before being evaluated in the environment.

\subsubsection{\pgame{} and optimization efficiency} While there was no significant difference in the \textit{final} QD scores of \cmamegatdes{} and \pgame{}, \cmamegatdes{} was less \textit{efficient} than \pgame{} in some environments. For instance, in QD Hopper, \pgame{} reached 1.5M QD score after 100k evaluations, but \cmamegatdes{} required 400k evaluations.

We can quantify optimization efficiency with \textit{QD score AUC}, the area under the curve (AUC) of the QD score plot. For a QD algorithm which executes $N$ iterations and evaluates $\lambda$ solutions per iteration, we define QD score AUC as a Riemann sum:
\ifdefined\modeWebsite
  QD score AUC $= \sum_{i=1}^N (\lambda * $ QD score at iteration $i)$
\else
\vspace{-0.5em}
\begin{align}
  \text{QD score AUC} = \sum_{i=1}^N (\lambda * \text{QD score at iteration $i$})
\end{align}
\fi
After computing QD score AUC, we ran statistical analysis similar to \sref{sec:analysis} and found \cmamegatdes{} had significantly lower QD score AUC than \pgame{} in QD Ant and QD Hopper. There was no significant difference in QD Half-Cheetah and QD Walker. As such, while \cmamegatdes{} obtained comparable final QD scores to \pgame{} in all tasks, it was less efficient at achieving those scores in QD Ant and QD Hopper.

\subsubsection{\mees{} and archive insertions} \label{sec:results-mees}

With one exception (\cmamegaes{} in QD Hopper), both \cmamega{} variants achieved significantly higher QD score than \mees{} in all environments. We attribute this result to the number of solutions each algorithm inserts into the archive. Each iteration, \mees{} evaluates 200 solutions (\apref{sec:algoparams}) but only inserts one into the archive, for a total of 5000 solutions inserted during each run. Given that each archive has at least 1000 cells, \mees{} has, on average, 5 opportunities to insert a solution that improves each cell. In contrast, the \cmamega{} variants have 100 times more insertions. Though the CMA-MEGA variants evaluate 200 solutions per iteration, they insert 100 of these into the archive. This totals to 500k insertions per run, allowing the CMA-MEGA variants to gradually improve archive cells.

\subsubsection{\mapelites{} and robustness} \label{sec:results-mapelites}

In most cases, both \cmamega{} variants had significantly higher QD score than \mapelites{} or no significant difference, but in QD Hopper, \mapelites{} achieved significantly higher QD score than \cmamegaes{}. However, we found that MAP-Elites solutions were less robust (see \apref{sec:results-mapelites-full}).

\subsubsection{\cmamega{} variants and gradient estimates} \label{sec:results-variants}

In QD Hopper and QD Walker, \cmamegatdes{} had significantly higher QD score than \cmamegaes{}. One potential explanation is that PyBullet Gym (and hence QDGym) augments rewards with reward shaping signals intended to promote optimal solutions for deep RL algorithms. In prior work \citep{pagliuca2020}, these signals led PPO \citep{schulman2017ppo} to train successful walking agents, while they led \openaies{} into local optima. For instance, \openaies{} trained agents which stood still so as to maximize only the reward signal for staying upright.

Due to these signals, TD3's objective gradient seems more useful than that of \openaies{} in QD Hopper and QD Walker. In fact, the algorithms which performed best in QD Hopper and QD Walker were ones that calculated objective gradients with TD3, i.e. \pgame{} and \cmamegatdes{}.

Prior work \citep{pagliuca2020} found that rewards could be tailored for ES, such that \openaies{} outperformed PPO. Extensions of our work could investigate whether there is a similar effect for QD algorithms, where tailoring the reward leads \cmamegaes{} to outperform \pgame{} and \cmamegatdes{}.

\section{Conclusion}

To extend DQD to RL settings, we adapted gradient approximations from actor-critic methods and ES. By integrating these approximations with \cmamega{}, we proposed two novel variants that we evaluated on four locomotion tasks from QDGym. \cmamegatdes{} performed comparably to the state-of-the-art \pgame{} in all tasks but was less efficient in two of the tasks. \cmamegaes{} performed comparably in two tasks.

Our results contrast prior work \citep{fontaine2021dqd} where \cmamega{} outperformed a baseline algorithm inspired by \pgame{} in QD benchmark domains. The difference seems to be that difficulty in the benchmarks arises from a hard-to-explore measure space, whereas difficulty in QDGym arises from an objective which requires rigorous optimization. As such, future work could formalize the notions of ``exploration difficulty'' of a measure space and ``optimization difficulty'' of an objective and evaluate algorithms in benchmarks that cover a spectrum of these metrics.

For practitioners looking to apply DQD in RL settings, we recommend estimating objective gradients with an off-policy actor-critic method such as TD3 instead of with an ES. Due to the difficulty of modern control benchmarks, it is important to efficiently optimize the objective --- TD3 benefits over ES since it can compute the objective gradient without further environment interaction. Furthermore, reward signals in these benchmarks are designed for deep RL methods, making TD3 gradients more useful than ES gradients.

By reducing \qdrl{} to DQD, we have decoupled \qdrl{} into DQD optimization and RL gradient approximations. In the future, we envision algorithms which benefit from advances in either more efficient DQD or more accurate RL gradient approximations.

\begin{acks}
  \ifModeWebsite{\lxAddClass{ltx2_acknowledgements}}
  \ifdefined\modeQuals
    This work is being submitted for the author's qualification exam at the University of Southern California. This work has also been accepted for publication at the 2022 Genetic and Evolutionary Computation Conference (GECCO '22). The author thanks Stefanos Nikolaidis (Chair), Satyandra K. Gupta, Haipeng Luo, Sven Koenig, and Gaurav Sukhatme for serving on his qualification committee. This work was done in collaboration with Matthew C. Fontaine, Julian Togelius, and Stefanos Nikolaidis.
  \else
    The authors thank the anonymous reviewers, Ya-Chuan Hsu, Heramb Nemlekar, and Gautam Salhotra for their invaluable feedback. This work was partially supported by the NSF NRI (\#1053128) and NSF GRFP (\#DGE-1842487). %
  \fi
\end{acks}

\ifdefined\modeWebsite
  \bibliographystyle{plain}
\else
  \bibliographystyle{ACM-Reference-Format}
\fi
\bibliography{references}

\appendix

\clearpage

\section{Helper Functions for \cmamega{} Variants} \label{sec:extra-code}

\begin{algorithm}
\caption{Helper function for updating the archive.}
\label{alg:update-archive}
\SetKwProg{UpdateArchive}{\textsc{Update\_Archive}}{:}{}
\UpdateArchive{$(\mathcal{A}, \vphi, f(\vphi), \vm(\vphi))$}{
  \tcp{$\mathcal{E}$ contains $\vphi_\mathcal{E}, f(\vphi_\mathcal{E}), \vm(\vphi_\mathcal{E})$}
  $\mathcal{E} \gets$ cell in $\mathcal{A}$ corresponding to $\vm$ \;
  \uIf{$\mathcal{E}$ is empty} {
    $\vphi_\mathcal{E}, f(\vphi_\mathcal{E}), \vm(\vphi_\mathcal{E}) \gets \vphi, f(\vphi), \vm(\vphi)$ \;
    \Return (\textsc{new\_cell}, $f(\vphi)$) \;
  }
  \uElseIf{$f(\vphi) > f(\vphi_\mathcal{E})$}{
    $\vphi_\mathcal{E}, f(\vphi_\mathcal{E}), \vm(\vphi_\mathcal{E}) \gets \vphi, f(\vphi), \vm(\vphi)$ \;
    \Return (\textsc{improve\_existing\_cell}, $f(\vphi) - f(\vphi_\mathcal{E})$)
  }
  \Else {
    \Return (\textsc{not\_added}, $f(\vphi) - f(\vphi_\mathcal{E})$)
  }
}
\end{algorithm}

\begin{algorithm}
\ifdefined\modeWebsite
  \caption{TD3 helper functions.}
\else
  \caption{TD3 helper functions. Adapted from \pgame{} \citep{nilsson2021pga} and TD3 \citep{fujimoto2018td3}.}
\fi
\label{alg:td3-helpers}

\SetKwProg{InitTD}{\textsc{Initialize\_TD3}}{:}{}
\InitTD{}{
  $\mathcal{B} \gets$ initialize\_replay\_buffer() \;
  \tcp{As done in the TD3 author implementation \citep{fujimoto2018td3}, we initialize these networks with the default PyTorch weights.}
  $Q_{\vtheta_1}, Q_{\vtheta_2}, \pi_{\vphi_q} \gets$ initialize\_networks() \;
  $Q_{\vtheta'_1}, Q_{\vtheta'_2}, \pi_{\vphi'_q} \gets Q_{\vtheta_1}, Q_{\vtheta_2}, \pi_{\vphi_q}$ \;
  \Return{$\mathcal{B}, Q_{\vtheta_1}, Q_{\vtheta_2}, \pi_{\vphi_q}, Q_{\vtheta'_1}, Q_{\vtheta'_2}, \pi_{\vphi'_q}$} \;
}

\SetKwProg{TDGrad}{\textsc{TD3\_Gradient}}{:}{}
\TDGrad{($\vphi, Q_{\vtheta_1}, \mathcal{B}$)}{
  Sample $n_{pg}$ transitions $(s_t, a_t, r(s_t,a_t), s_{t+1})$ from $\mathcal{B}$ \;
  $\nabla_\vphi J(\vphi) = \frac{1}{n_{pg}} \sum \nabla_\vphi \pi_\vphi(s_t) \nabla_a Q_{\vtheta_1}(s_t,a) |_{a = \pi_\vphi(s_t)}$ \;
  \Return{$\nabla_\vphi J(\vphi)$} \;
}

\SetKwProg{TrainTD}{\textsc{Train\_TD3}}{:}{}
\TrainTD{$(Q_{\vtheta_1}, Q_{\vtheta_2}, \pi_{\vphi_q} Q_{\vtheta'_1}, Q_{\vtheta'_2}, \pi_{\vphi'_q}, \mathcal{B})$}{
  \tcp{Trains the critic and the greedy actor.}
  \For{$i \gets 1..n_{crit}$} {
    Sample $n_q$ transitions $(s_t, a_t, r(s_t,a_t), s_{t+1})$ from $\mathcal{B}$ \;
    \tcp{Sample smoothing noise.}
    $\epsilon \sim$ clip$(\mathcal{N}(0, \sigma_p), -c_{clip}, c_{clip})$ \;
    $y = r(s_t, a_t) + \gamma \min_{i=1,2}Q_{\vtheta'_i}(s_{t+1},\pi_{\vphi'_q}(s_{t+1})+\epsilon)$ \;
    \tcp{Update critics.}
    $\vtheta_i \gets \argmin_{\vtheta_i} \frac{1}{n_q} \sum(y - Q_{\vtheta_i}(s_t,a_t))^2$ \;
    \If{$t \mod d = 0$}{
      \tcp{Update greedy actor.}
      $\nabla_{\vphi_q} J(\vphi_q) = \frac{1}{n_q} \sum \nabla_{\vphi_q} \pi_{\vphi_q}(s_t) \nabla_a Q_{\vtheta_1}(s_t,a) |_{a = \pi_{\vphi_q}(s_t)}$ \;
      \tcp{Update targets.}
      $\vtheta'_i \gets \tau \vtheta_i + (1 - \tau)\vtheta'_i$ \;
      $\vphi'_q \gets \tau \vphi_q + (1 - \tau)\vphi'_q$ \;
    }
  }
}
\end{algorithm}

\begin{algorithm}
\caption{Helper function for estimating objective and measure gradients with the gradient estimate from \openaies{}. \ifdefined\modeWebsite\else This implementation differs from \eref{eq:openaiesgrad} since it includes mirror sampling and rank normalization.\fi}
\label{alg:estimate-es-gradient}
\SetKwProg{ESGradients}{\textsc{ES\_Gradients}}{:}{}
\ESGradients{$(\vphi, \lambda_{es}, \sigma_e)$}{
  \tcp{Mirror sampling - divide $\lambda_{es}$ by 2.}
  \For{$i \gets 1..\frac{\lambda_{es}}{2}$} {
    $\vepsilon_i \sim \gN(\mathbf{0}, \mI)$ \;
    $\vx_i \gets \vphi + \sigma_e\vepsilon_i$ \;
    $f(\vx_i), \vm(\vx_i) \gets evaluate(\vx_i)$ \;
    $\vx'_i \gets \vphi - \sigma_e\vepsilon_i$ \tcp*{$\vx'_i$ reflects $\vx_i$.}
    $f(\vx'_i), \vm(\vx'_i) \gets evaluate(\vx_i)$ \;
  }

  \For{$j \gets 0..k$} {
    \uIf{$j = 0$} {
      $L \gets$ all $\vx_i$ and $\vx'_i$, sorted by $f$
    } \Else {
      $L \gets$ all $\vx_i$ and $\vx'_i$, sorted by $m_j$
    }
    \tcp{Rank normalization.}
    $R \gets$ rank (index) of every $\vx_i$ in $L$ \;
    $R' \gets$ rank (index) of every $\vx'_i$ in $L$ \;
    \tcp{Ranks should be normalized over both lists combined ($R \| R'$) rather than in each list separately.}
    Normalize ranks in $R \| R'$ to $[-0.5, 0.5]$ \;

    \tcp{Estimate gradient.}
    $\vnabla \gets \frac{1}{\frac{\lambda_{es}}{2}\sigma_e} \sum_{i=1}^{\frac{\lambda_{es}}{2}} \vepsilon_i(R_i - R'_i)$

    \uIf{$j = 0$} {
      $\vnabla f(\vphi) \gets \vnabla$
    } \Else {
      $\vnabla {m_j}(\vphi) \gets \vnabla$
    }
  }

  \Return {$\vnabla f(\vphi), \vnabla \vm(\vphi)$}
}
\end{algorithm}

\clearpage

\section{Algorithm Hyperparameters} \label{sec:algoparams}

Here we list parameters for each algorithm in our experiments. Refer to \sref{sec:envparams} for parameters of the neural network policy and the archive. All algorithms are allocated 1,000,000 evaluations total.

\begin{table}[ht]
\caption{\cmamegaes{} and \cmamegatdes{} hyperparameters. $n_{pg}$ and $n_{crit}$ are only applicable in \cmamegatdes{}. $n_{pg}$ here is analogous to $n_{pg}$ in \pgame{}, but we make it much larger here (65,536 vs. 256) to improve the accuracy of the gradient estimate. It is important to obtain a more accurate gradient estimate since we only compute one gradient per iteration instead of taking gradient steps on multiple solutions.}
\label{table:params-cmamegaes}
\begin{center}
\begin{tabular}{C{0.6in} L{1.7in} C{0.6in}}
\toprule
Parameter & Description & Value \\
\midrule
$N$ & Iterations = 1,000,000 / ($\lambda + \lambda_{es}$) & 5,000 \\
$\lambda$ & Batch size & 100 \\
$\sigma_g$ & Initial \cmaes{} step size & 1.0 \\
$\eta$ & Gradient ascent learning rate & 1.0 \\
$\lambda_{es}$ & ES batch size & 100 \\
$\sigma_e$ & ES noise standard deviation & 0.02 \\
$n_{pg}$ & TD3 gradient estimate batch size & 65,536 \\
$n_{crit}$ & TD3 critic training steps & 600 \\
\bottomrule
\end{tabular}
\end{center}
\end{table}

\begin{table}[ht]
\caption{\pgame{} hyperparameters.}
\label{table:params-pgame}
\begin{center}
\begin{tabular}{C{0.6in} L{1.7in} C{0.6in}}
\toprule
Parameter & Description & Value \\
\midrule
$N$ & Iterations = 1,000,000 / $\lambda$ & 10,000 \\
$\lambda$ & Batch size & 100 \\
$n_{evo}$ & Variation operators split & $0.5\lambda = 50$ \\
$n_{grad}$ & PG variation steps & 10 \\
$\alpha_{grad}$ & PG variation learning rate (for Adam) & 0.001 \\
$n_{pg}$ & PG variation batch size & 256 \\
$n_{crit}$ & TD3 critic training steps & 300 \\
$\sigma_1$ & GA variation 1 & 0.005 \\
$\sigma_2$ & GA variation 2 & 0.05 \\
$G$ & Random initial solutions & 100 \\
\bottomrule
\end{tabular}
\end{center}
\end{table}

\begin{table}[ht]
\caption{\mees{} hyperparameters. We adopt the explore-exploit variant.}
\label{table:params-mees}
\begin{center}
\begin{tabular}{C{0.6in} L{1.7in} C{0.6in}}
\toprule
Parameter & Description & Value \\
\midrule
$N$ & Iterations = 1,000,000 / $\lambda$ & 5,000 \\
$\lambda$ & Batch size & 200 \\
$\sigma$ & ES noise standard deviation & 0.02 \\
$n_{optim\_gens}$ & Consecutive generations to optimize a solution & 10 \\
$\alpha$ & Learning rate for Adam & 0.01 \\
$\alpha_2$ & L2 coefficient for Adam & 0.005 \\
$k$ & Nearest neighbors for novelty calculation & 10 \\
\bottomrule
\end{tabular}
\end{center}
\end{table}

\begin{table}[ht]
\caption{\mapelites{} hyperparameters. We describe \mapelites{} in \sref{sec:mapelites}.}
\label{table:params-mapelites}
\begin{center}
\begin{tabular}{C{0.6in} L{1.7in} C{0.6in}}
\toprule
Parameter & Description & Value \\
\midrule
$N$ & Iterations = 1,000,000 / $\lambda$ & 10,000 \\
$\lambda$ & Batch size & 100 \\
$\sigma$ & Gaussian noise standard deviation & 0.02 \\
\bottomrule
\end{tabular}
\end{center}
\end{table}

\begin{table}[ht]
\caption{TD3 hyperparameters common to \cmamegatdes{} and \pgame{}, which both train a TD3 instance. Furthermore, though we record the objective with $\gamma = 1$ (\sref{sec:metrics}), TD3 still executes with $\gamma < 1$.}
\label{table:params-td3}
\begin{center}
\begin{tabular}{C{0.6in} L{1.7in} C{0.6in}}
\toprule
Parameter & Description & Value \\
\midrule
--- & Critic layer sizes  & $[256,256,1]$ \\
$\alpha_{crit}$ & Critic learning rate (for Adam) & 3e-4 \\
$n_q$ & Critic training batch size & 256 \\
$|\mathcal{B}|$ & Max replay buffer size & 1,000,000 \\
$\gamma$ & Discount factor & 0.99 \\
$\tau$ & Target network update rate & 0.005 \\
$d$ & Target network update frequency & 2 \\
$\sigma_p$ & Smoothing noise standard deviation & 0.2 \\
$c_{clip}$ & Smoothing noise clip & 0.5 \\
\bottomrule
\end{tabular}
\end{center}
\end{table}

\clearpage

\section{Environment Details} \label{sec:environment_details}

\ifdefined\modeQuals
\begin{table}[t]
\else
\begin{table}[ht]
\fi

\ifModeWebsite{\lxAddClass{mount_envs}}
  \caption{QDGym environments details. We list the dimensions of the state space ($|\mathcal{S}|$) and action space ($|\mathcal{U}|$), number of neural network parameters, number of measures $|\mathcal{X}|$, archive grid dimensions (number of cells along each dimension), total archive grid cells, and min and max objectives (\sref{sec:metrics}).}
\label{table:envs}
\begin{center}
\begin{tabular}{lrrrr}
\toprule

{} &
\multicolumn{1}{p{0.12\linewidth}}{\centering QD\\Ant}
   &
\multicolumn{1}{p{0.21\linewidth}}{\centering QD\\Half-Cheetah}
   &
\multicolumn{1}{p{0.12\linewidth}}{\centering QD\\Hopper}
   &
\multicolumn{1}{p{0.12\linewidth}}{\centering QD\\Walker}
\vspace{5pt}
\\

{} &
\multicolumn{1}{c}{\includegraphics[width=0.12\linewidth]{imgs/qdgym_env_imgs/ant.png}}
   &
\multicolumn{1}{c}{\includegraphics[width=0.12\linewidth]{imgs/qdgym_env_imgs/half_cheetah.png}}
   &
\multicolumn{1}{c}{\includegraphics[width=0.036\linewidth]{imgs/qdgym_env_imgs/hopper.png}}
   &
\multicolumn{1}{c}{\includegraphics[width=0.05\linewidth]{imgs/qdgym_env_imgs/walker.png}}
\\

\midrule
$|\mathcal{S}|$ & 28 & 26 & 15 & 22 \\
$|\mathcal{U}|$ & 8 & 6 & 3 & 6 \\
Parameters & 21,256 & 20,742 & 18,947 & 20,230 \\
$|\mathcal{X}|$ & 4 & 2 & 1 & 2 \\
Archive dim & [6,6,6,6] & [32,32] & [1024] & [32,32] \\
Grid cells & 1,296 & 1,024 & 1,024 & 1,024 \\
Min objective & -374.70 & -2,797.52 & -362.09 & -67.17 \\
Max objective & 2,500.00 & 3,000.00 & 2,500.00 & 2,500.00 \\
\bottomrule
\end{tabular}
\end{center}
\end{table}

\tref{table:envs} lists all environment details. The measures in QDGym are the proportions of time that each foot contacts the ground. In each environment, the feet are ordered as follows:

\begin{itemize}
  \item \textbf{QD Ant:} front left foot, front right foot, back left foot, back right foot
  \item \textbf{QD Half-Cheetah:} front foot, back foot
  \item \textbf{QD Hopper:} single foot
  \item \textbf{QD Walker:} right foot, left foot
\end{itemize}

\section{\mapelites{} and robustness} \label{sec:results-mapelites-full}

In most cases, both \cmamega{} variants had significantly higher QD score than \mapelites{} or no significant difference, but in QD Hopper, \mapelites{} achieved significantly higher QD score than \cmamegaes{}. However, when we visualized solutions found by \mapelites{}, their performance was lower than the performance recorded in the archive. The best \mapelites{} solution in QD Hopper hopped forward a few steps and fell down, despite recording an excellent performance of \ifdefined\modeWebsite 2,648.31.\else 2,648.31 (see supplemental videos).\fi

One explanation for this behavior is that since we only evaluate solutions for one episode before inserting into the archive, a solution with noisy performance may be inserted because of a single high-performing episode, even if it performs poorly on average. Prior work \citep{nilsson2021pga} has also encountered this issue when running \mapelites{} with a directional variation operator \citep{vassiliades2018line} in QDGym, and has suggested measuring \textit{robustness} as a proxy for how much noise is present in an archive's solutions. Robustness is defined as the difference between the mean performance of the solution over $n$ episodes (we use $n=10$) and the performance recorded in the archive. The larger (more negative) this difference, the more noisy and less robust the solution.

To compare the robustness of the solutions output by the \cmamega{} variants and \mapelites{}, we computed \textit{mean elite robustness}, the average robustness of all elites in each experiment's final archive. We then ran statistical analysis similar to \sref{sec:analysis}. In all environments, both \cmamegaes{} and \cmamegatdes{} had significantly higher mean elite robustness than \mapelites{} (\apref{sec:finalmetrics} \& \ref{sec:fullstats}). Overall, though \mapelites{} achieves high QD score, its solutions are less robust.

\section{Final Metrics} \label{sec:finalmetrics}

Tables \ref{table:qd-score}-\ref{table:runtime-hours} show the QD score (\sref{sec:metrics}), QD score AUC (\sref{sec:results-pgame}), archive coverage (\sref{sec:metrics}), best performance (\sref{sec:metrics}), mean elite robustness (\sref{sec:results-mapelites}), and runtime in hours 
for all algorithms in all environments. The tables show the value of each metric after 1 million evaluations, averaged over 5 trials. Due to its magnitude, QD score AUC is expressed as a multiple of $10^{12}$.

Though \cmamegatdes{} and \pgame{} perform best overall, they rely on specialized hardware (a GPU) and require the most computation. As shown in \tref{table:runtime-hours}, the TD3 training in these algorithms leads to long runtimes. When runtime is dominated by the algorithm itself (as opposed to solution evaluations), \cmamegaes{} offers a viable alternative that may achieve reasonable performance.

\clearpage

\begin{table*}[t]
\caption{QD Score}
\label{table:qd-score}
\begin{center}
\begin{tabular}{l R{0.9in} R{0.9in} R{0.9in} R{0.9in}}
\toprule
{} &        QD Ant & QD Half-Cheetah &     QD Hopper &     QD Walker \\
\midrule
CMA-MEGA (ES)      &  1,649,846.69 &    4,489,327.04 &  1,016,897.48 &    371,804.19 \\
CMA-MEGA (TD3, ES) &  1,479,725.62 &    4,612,926.99 &  1,857,671.12 &  1,437,319.62 \\
PGA-MAP-Elites     &  1,674,374.81 &    4,758,921.89 &  2,068,953.54 &  1,480,443.84 \\
ME-ES              &    539,742.08 &    2,296,974.58 &    791,954.55 &    105,320.97 \\
MAP-Elites         &  1,418,306.56 &    4,175,704.19 &  1,835,703.73 &    447,737.90 \\
\bottomrule
\end{tabular}
\end{center}
\end{table*}

\begin{table*}[t]
\caption{QD Score AUC (multiple of $10^{12}$)}
\label{table:qd-score-auc}
\begin{center}
\begin{tabular}{l R{0.9in} R{0.9in} R{0.9in} R{0.9in}}
\toprule
{} & QD Ant & QD Half-Cheetah & QD Hopper & QD Walker \\
\midrule
CMA-MEGA (ES)      &   1.31 &            3.96 &      0.74 &      0.28 \\
CMA-MEGA (TD3, ES) &   1.14 &            3.97 &      1.39 &      1.01 \\
PGA-MAP-Elites     &   1.39 &            4.39 &      1.81 &      1.04 \\
ME-ES              &   0.35 &            1.57 &      0.49 &      0.07 \\
MAP-Elites         &   1.18 &            3.78 &      1.34 &      0.35 \\
\bottomrule
\end{tabular}
\end{center}
\end{table*}

\begin{table*}[t]
\caption{Archive Coverage}
\label{table:archive-coverage}
\begin{center}
\begin{tabular}{l R{0.9in} R{0.9in} R{0.9in} R{0.9in}}
\toprule
{} & QD Ant & QD Half-Cheetah & QD Hopper & QD Walker \\
\midrule
CMA-MEGA (ES)      &   0.96 &            1.00 &      0.97 &      1.00 \\
CMA-MEGA (TD3, ES) &   0.97 &            1.00 &      0.98 &      1.00 \\
PGA-MAP-Elites     &   0.96 &            1.00 &      0.97 &      0.99 \\
ME-ES              &   0.63 &            0.95 &      0.74 &      0.86 \\
MAP-Elites         &   0.98 &            1.00 &      0.98 &      1.00 \\
\bottomrule
\end{tabular}
\end{center}
\end{table*}

\begin{table*}[t]
\caption{Best Performance}
\label{table:best-performance}
\begin{center}
\begin{tabular}{l R{0.9in} R{0.9in} R{0.9in} R{0.9in}}
\toprule
{} &    QD Ant & QD Half-Cheetah & QD Hopper & QD Walker \\
\midrule
CMA-MEGA (ES)      &  2,213.06 &        2,265.73 &  1,441.00 &    940.50 \\
CMA-MEGA (TD3, ES) &  2,482.83 &        2,486.10 &  2,597.87 &  2,302.31 \\
PGA-MAP-Elites     &  2,843.86 &        2,746.98 &  2,884.08 &  2,619.17 \\
ME-ES              &  2,515.20 &        1,911.33 &  2,642.30 &  1,025.74 \\
MAP-Elites         &  1,506.97 &        1,822.88 &  2,602.94 &    989.31 \\
\bottomrule
\end{tabular}
\end{center}
\end{table*}

\begin{table*}[t]
\caption{Mean Elite Robustness}
\label{table:mean-elite-robustness}
\begin{center}
\begin{tabular}{l R{0.9in} R{0.9in} R{0.9in} R{0.9in}}
\toprule
{} &   QD Ant & QD Half-Cheetah & QD Hopper & QD Walker \\
\midrule
CMA-MEGA (ES)      &   -51.62 &         -105.81 &   -187.44 &    -86.45 \\
CMA-MEGA (TD3, ES) &   -48.91 &          -80.78 &   -273.68 &    -97.40 \\
PGA-MAP-Elites     &    -4.16 &          -92.38 &   -435.45 &    -74.26 \\
ME-ES              &    77.76 &         -645.40 &   -631.32 &      2.05 \\
MAP-Elites         &  -109.42 &         -338.78 &   -509.21 &   -186.14 \\
\bottomrule
\end{tabular}
\end{center}
\end{table*}

\begin{table*}[t]
\caption{Runtime (Hours)}
\label{table:runtime-hours}
\begin{center}
\begin{tabular}{l R{0.9in} R{0.9in} R{0.9in} R{0.9in}}
\toprule
{} & QD Ant & QD Half-Cheetah & QD Hopper & QD Walker \\
\midrule
CMA-MEGA (ES)      &   7.40 &            7.24 &      3.84 &      3.52 \\
CMA-MEGA (TD3, ES) &  16.26 &           22.79 &     13.43 &     13.01 \\
PGA-MAP-Elites     &  19.99 &           19.75 &     12.65 &     12.86 \\
ME-ES              &   8.92 &           10.25 &      4.04 &      4.12 \\
MAP-Elites         &   7.43 &            7.37 &      4.59 &      5.72 \\
\bottomrule
\end{tabular}
\end{center}
\end{table*}

\clearpage

\section{Full Statistical Analysis} \label{sec:fullstats}

To compare a metric such as QD score between two or more algorithms across all four QDGym environments, we performed a two-way ANOVA where environment and algorithm were the independent variables and the metric was the dependent variable. When there was a significant interaction effect (note that all of our analyses found significant interaction effects), we followed up this ANOVA with a simple main effects analysis in each environment. Finally, we ran pairwise comparisons (two-sided t-tests) to determine which algorithms had a significant difference on the metric. We applied Bonferroni corrections within each environment / simple main effect. For example, in \tref{table:normalized-qd-score-ttests-0} we compared \cmamegaes{} with three algorithms in each environment, so we applied a Bonferroni correction with $n=3$.

This section lists the ANOVA and pairwise comparison results for each of our analyses. We have \textbf{bolded} all significant $p$-values, where significance is determined at the $\alpha = 0.05$ threshold. For pairwise comparisons, some $p$-values are marked as ``1'' because the Bonferroni correction caused the $p$-value to exceed 1. $p$-values less than 0.001 have been marked as ``\textbf{< 0.001}''.

\subsection{QD Score Analysis (\sref{sec:analysis})}

To test the hypotheses we defined in \sref{sec:experimentdesign}, we performed a two-way ANOVA for QD scores. Since the ANOVA requires scores in all environments to have the same scale, we normalized the QD score in all environments by dividing by the maximum QD score, defined in \sref{sec:analysis} as \textit{grid cells} * (\textit{max objective} - \textit{min objective}). The results of the ANOVA were as follows:

\begin{itemize}
\item Interaction effect: $F(12, 80) = 16.82, \mathbf{p < 0.001}$
\item Simple main effects:
  \begin{itemize}
  \item QD Ant: $F(4, 80) = 23.87, \mathbf{p < 0.001}$
  \item QD Half-Cheetah: $F(4, 80) = 44.15, \mathbf{p < 0.001}$
  \item QD Hopper: $F(4, 80) = 57.35, \mathbf{p < 0.001}$
  \item QD Walker: $F(4, 80) = 90.84, \mathbf{p < 0.001}$
  \end{itemize}
\end{itemize}

Since the ANOVA showed a significant interaction effect and significant simple main effects, we performed pairwise comparisons for each hypothesis (Tables \ref{table:normalized-qd-score-ttests-0}-\ref{table:normalized-qd-score-ttests-2}).

\subsection{QD Score AUC Analysis (\sref{sec:results-pgame})}

In this followup analysis, we hypothesized that \pgame{} would have greater QD score AUC than \cmamegaes{} and \cmamegatdes{}. Thus, we performed a two-way ANOVA which compared QD score AUC for \pgame{}, \cmamegaes{}, and \cmamegatdes{}. As we did for QD score, we normalized QD score AUC by the maximum QD score. The ANOVA results were as follows:

\begin{itemize}
\item Interaction effect: $F(12, 80) = 17.55, \mathbf{p < 0.001}$
\item Simple main effects:
  \begin{itemize}
  \item QD Ant: $F(4, 80) = 31.77, \mathbf{p < 0.001}$
  \item QD Half-Cheetah: $F(4, 80) = 89.38, \mathbf{p < 0.001}$
  \item QD Hopper: $F(4, 80) = 82.34, \mathbf{p < 0.001}$
  \item QD Walker: $F(4, 80) = 71.64, \mathbf{p < 0.001}$
  \end{itemize}
\end{itemize}

As the interaction and simple main effects were significant, we performed pairwise comparisons (\tref{table:normalized-qd-score-auc-ttests-0}).

\subsection{Mean Elite Robustness Analysis (\sref{sec:results-mapelites})}

In this followup analysis, we hypothesized that \mapelites{} would have lower mean elite robustness than \cmamegaes{} and \cmamegatdes{}. Thus, we performed a two-way ANOVA which compared mean elite robustness for \mapelites{}, \cmamegaes{}, and \cmamegatdes{}. We normalized by the score range, i.e. \textit{max objective} - \textit{min objective}. The ANOVA results were as follows:

\begin{itemize}
\item Interaction effect: $F(12, 80) = 8.75, \mathbf{p < 0.001}$
\item Simple main effects:
  \begin{itemize}
  \item QD Ant: $F(4, 80) = 3.17, \mathbf{p = 0.018}$
  \item QD Half-Cheetah: $F(4, 80) = 9.60, \mathbf{p < 0.001}$
  \item QD Hopper: $F(4, 80) = 21.07, \mathbf{p < 0.001}$
  \item QD Walker: $F(4, 80) = 3.70, \mathbf{p = 0.008}$
  \end{itemize}
\end{itemize}

As the interaction and simple main effects were significant, we performed pairwise comparisons (\tref{table:normalized-mean-elite-robustness-ttests-0}).

\ifdefined\modeQuals
    \begin{landscape}
\fi

\begin{table*}[t]
\caption{H1 - Comparing QD score between CMA-MEGA (ES) and baselines}
\label{table:normalized-qd-score-ttests-0}
\begin{center}
\begin{tabular}{ L{1.2 in} L{1.2 in} R{0.9in} R{0.9in} R{0.9in} R{0.9in}}
\toprule
              &            &            QD Ant &   QD Half-Cheetah &       QD Hopper &         QD Walker \\
Algorithm 1 & Algorithm 2 &                   &                   &                 &                   \\
\midrule
\multirow{3}{*}{CMA-MEGA (ES)} & PGA-MAP-Elites &                 1 &             0.733 &  \textbf{0.003} &  \textbf{< 0.001} \\
              & ME-ES &  \textbf{< 0.001} &  \textbf{< 0.001} &           0.841 &  \textbf{< 0.001} \\
              & MAP-Elites &             0.254 &             0.215 &  \textbf{0.007} &             0.108 \\
\bottomrule
\end{tabular}
\end{center}
\end{table*}

\begin{table*}[t]
\caption{H2 - Comparing QD score between CMA-MEGA (TD3, ES) and baselines}
\label{table:normalized-qd-score-ttests-1}
\begin{center}
\begin{tabular}{ L{1.2 in} L{1.2 in} R{0.9in} R{0.9in} R{0.9in} R{0.9in}}
\toprule
                   &            &            QD Ant &   QD Half-Cheetah &         QD Hopper &         QD Walker \\
Algorithm 1 & Algorithm 2 &                   &                   &                   &                   \\
\midrule
\multirow{3}{*}{CMA-MEGA (TD3, ES)} & PGA-MAP-Elites &             0.093 &                 1 &             0.726 &                 1 \\
                   & ME-ES &  \textbf{< 0.001} &  \textbf{< 0.001} &  \textbf{< 0.001} &  \textbf{< 0.001} \\
                   & MAP-Elites &                 1 &    \textbf{0.010} &                 1 &  \textbf{< 0.001} \\
\bottomrule
\end{tabular}
\end{center}
\end{table*}

\begin{table*}[t]
\caption{H3 - Comparing QD score between CMA-MEGA (ES) and CMA-MEGA (TD3, ES)}
\label{table:normalized-qd-score-ttests-2}
\begin{center}
\begin{tabular}{ L{1.2 in} L{1.2 in} R{0.9in} R{0.9in} R{0.9in} R{0.9in}}
\toprule
              &                    & QD Ant & QD Half-Cheetah &       QD Hopper &         QD Walker \\
Algorithm 1 & Algorithm 2 &        &                 &                 &                   \\
\midrule
CMA-MEGA (ES) & CMA-MEGA (TD3, ES) &  0.250 &           0.511 &  \textbf{0.006} &  \textbf{< 0.001} \\
\bottomrule
\end{tabular}
\end{center}
\end{table*}

\begin{table*}[t]
\caption{Comparing QD score AUC between PGA-ME and CMA-MEGA variants}
\label{table:normalized-qd-score-auc-ttests-0}
\begin{center}
\begin{tabular}{ L{1.2 in} L{1.2 in} R{0.9in} R{0.9in} R{0.9in} R{0.9in}}
\toprule
               &                    &          QD Ant & QD Half-Cheetah &         QD Hopper &         QD Walker \\
Algorithm 1 & Algorithm 2 &                 &                 &                   &                   \\
\midrule
\multirow{2}{*}{PGA-MAP-Elites} & CMA-MEGA (ES) &           0.734 &           0.255 &  \textbf{< 0.001} &  \textbf{< 0.001} \\
               & CMA-MEGA (TD3, ES) &  \textbf{0.020} &           0.111 &    \textbf{0.003} &                 1 \\
\bottomrule
\end{tabular}
\end{center}
\end{table*}

\begin{table*}[t]
\caption{Comparing mean elite robustness between MAP-Elites and CMA-MEGA variants}
\label{table:normalized-mean-elite-robustness-ttests-0}
\begin{center}
\begin{tabular}{ L{1.2 in} L{1.2 in} R{0.9in} R{0.9in} R{0.9in} R{0.9in}}
\toprule
           &                    &            QD Ant &   QD Half-Cheetah &       QD Hopper &         QD Walker \\
Algorithm 1 & Algorithm 2 &                   &                   &                 &                   \\
\midrule
\multirow{2}{*}{MAP-Elites} & CMA-MEGA (ES) &  \textbf{< 0.001} &  \textbf{< 0.001} &  \textbf{0.030} &    \textbf{0.003} \\
           & CMA-MEGA (TD3, ES) &  \textbf{< 0.001} &  \textbf{< 0.001} &  \textbf{0.013} &  \textbf{< 0.001} \\
\bottomrule
\end{tabular}
\end{center}
\end{table*}

\ifdefined\modeQuals
    \end{landscape}
\fi

\clearpage

\section{Archive Visualizations} \label{sec:archives}

We visualize ``median'' archives in \fref{fig:heatmaps} and \fref{fig:histograms}. To determine these median archives, we selected the trial which achieved the median QD score out of the 5 trials of each algorithm in each environment. \fref{fig:heatmaps} visualizes heatmaps of median archives in QD Half-Cheetah and QD Walker, while \fref{fig:histograms} shows the distribution (histogram) of objective values for median archives in all environments.\ifModeWebsiteOff{ Refer to the supplemental material for videos of how these figures develop across iterations.}

\clearpage %

\begin{figure*}[t]
\ifModeWebsite{\lxAddClass{mount_heatmaps}}
\begin{center}
\includegraphics[width=\linewidth]{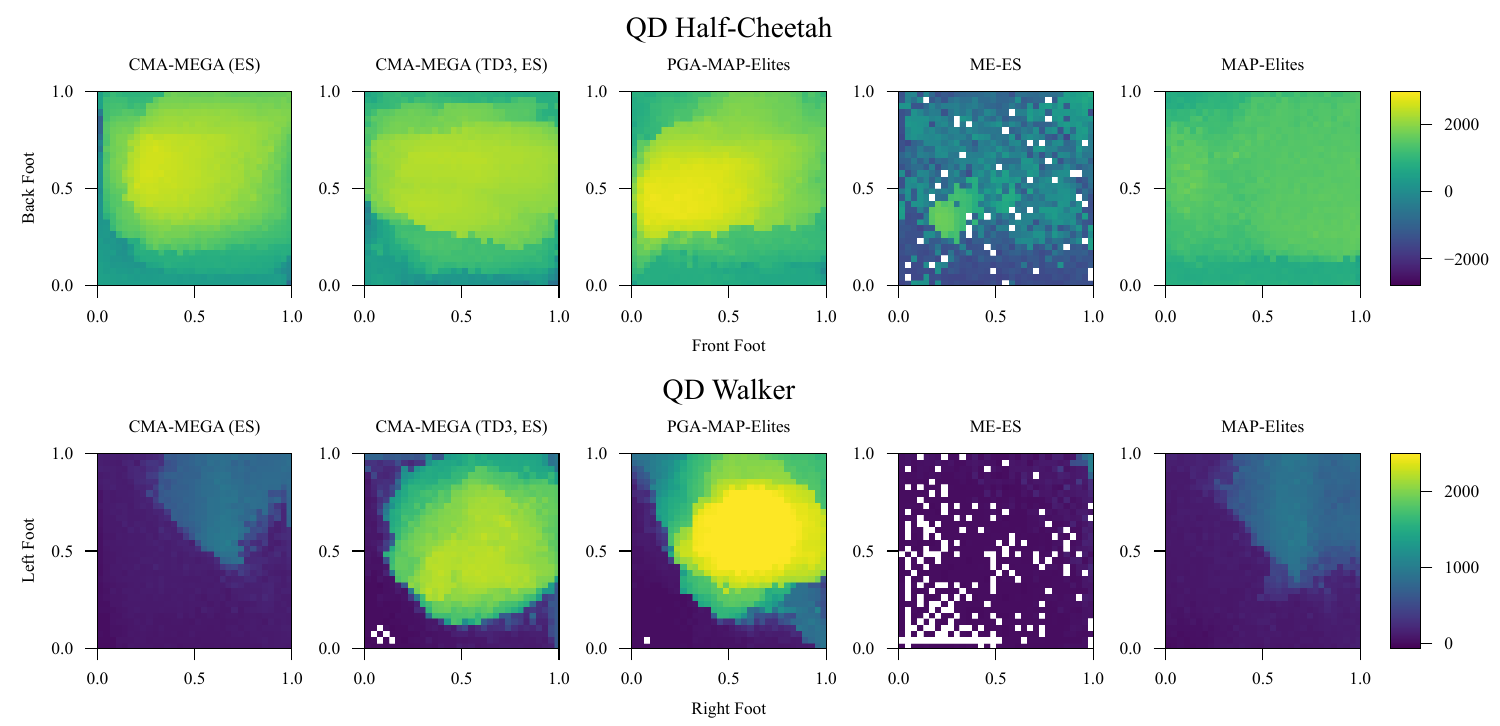}
\end{center}
\caption{Archive heatmaps from the median trial (in terms of QD score) of each algorithm in QD Half-Cheetah and QD Walker. The colorbar for each environment ranges from the minimum to maximum objective stated in \tref{table:envs}. The archive in both environments is a $32 \times 32$ grid. Currently, we are unable to plot heatmaps for QD Ant and QD Hopper because their archives are not 2D.\ifModeWebsiteOff{ Refer to the supplemental material for a video of how these archives develop across iterations.}
\\
\\
These heatmaps have several notable features. First, we can see that \mapelites{} primarily discovers low-performing solutions. Second, from looking at the heatmap videos, we can see that \pgame{} gradually improves the entire archive ``all at once'' --- this happens because \pgame{} samples solutions uniformly from the archive and applies variations to them, so the entire archive appears to improve simultaneously. Finally, again based on the heatmap videos, we see that the \cmamega{} variants improve the archive with ``paintbrush strokes.'' This happens because the \cmamega{} variants gradually move the solution point $\vphi^*$ around the archive while generating solutions around it.}
\label{fig:heatmaps}
\Description[Fully described in the text]{}
\end{figure*}

\begin{figure*}[t]
\ifModeWebsite{\lxAddClass{mount_histograms}}
\begin{center}
\includegraphics[width=\linewidth]{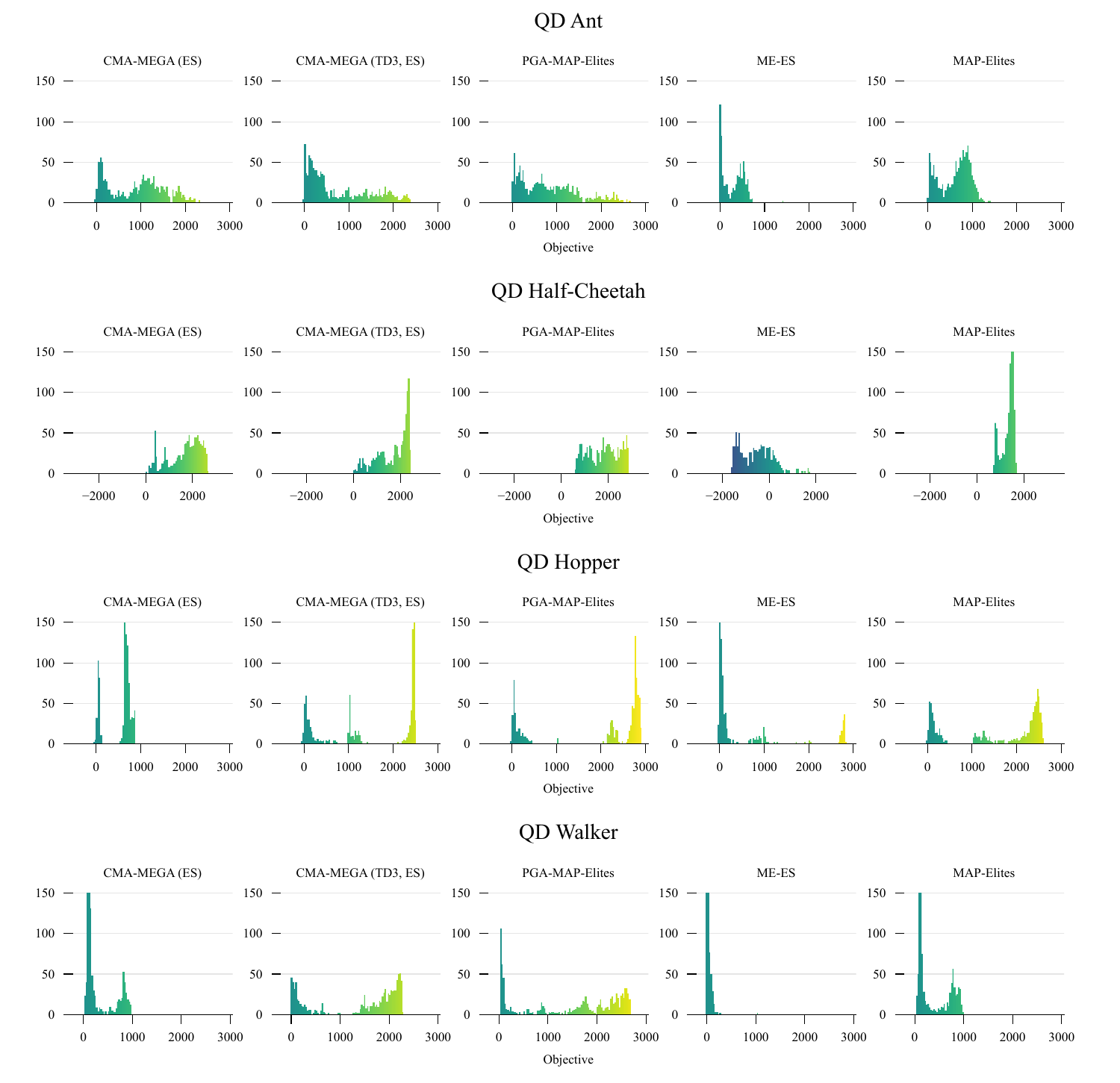}
\end{center}
\caption{Distribution (histogram) of objective values in archives from the median trial (in terms of QD score) of each algorithm in each environment. In each plot, the x-axis is bounded on the left by the minimum objective and on the right by the maximum objective plus 400, as some solutions exceed the maximum objective in \tref{table:envs}. Note that in some plots, the number of items overflows the y-axis bounds (e.g. \mees{} in QD Walker).\ifModeWebsiteOff{ Refer to the supplemental material for a video of how these distributions develop across iterations.}}
\label{fig:histograms}
\Description[Fully described in the text]{}
\end{figure*}

\end{document}